\pdfoutput=1

\documentclass[11pt]{article}

\usepackage[]{ACL2023}

\usepackage{times}
\usepackage{latexsym}

\usepackage[T1]{fontenc}

\usepackage[utf8]{inputenc}

\usepackage{microtype}

\usepackage{inconsolata}

\usepackage{booktabs}       
\usepackage{amsfonts}       
\usepackage{nicefrac}       
\usepackage{microtype}      
\usepackage{xcolor}         
\usepackage{overpic}

\usepackage{multirow}
\usepackage{mathtools}
\usepackage{subcaption}
\usepackage{soul}

\title{Reversed Attention: On The Gradient Descent Of Attention Layers In GPT}

\author{Shahar Katz ~~~~~~ ~~~~~~ Lior Wolf\\
Blavatnik School of Computer Science, Tel Aviv University\\
\small{\texttt{\{shaharkatz3@mail,wolf@cs\}.tau.ac.il}}
}

\begin{document}
\maketitle
\begin{abstract}
The success of Transformer-based Language Models (LMs) stems from their attention mechanism. 
While this mechanism has been extensively studied in explainability research, particularly through the attention values obtained during the forward pass of LMs, the backward pass of attention has been largely overlooked.
In this work, we study the mathematics of the backward pass of attention, revealing that it implicitly calculates an attention matrix we refer to as ``Reversed Attention''.
We examine the properties of Reversed Attention and demonstrate its ability to elucidate the models' behavior and edit dynamics.
In an experimental setup, we showcase the ability of Reversed Attention to directly alter the forward pass of attention, without modifying the model's weights, using a novel method called ``attention patching''.
In addition to enhancing the comprehension of how LM configure attention layers during backpropagation, Reversed Attention maps contribute to a more interpretable backward pass.
Our code will be available at: \url{https://github.com/shacharKZ/Reversed-Attention}  .
\end{abstract}

\begin{figure}
  \vspace{-1cm}
  \centering
  \rule[-.5cm]{0cm}{4cm}
  \includegraphics[width=0.95\columnwidth]{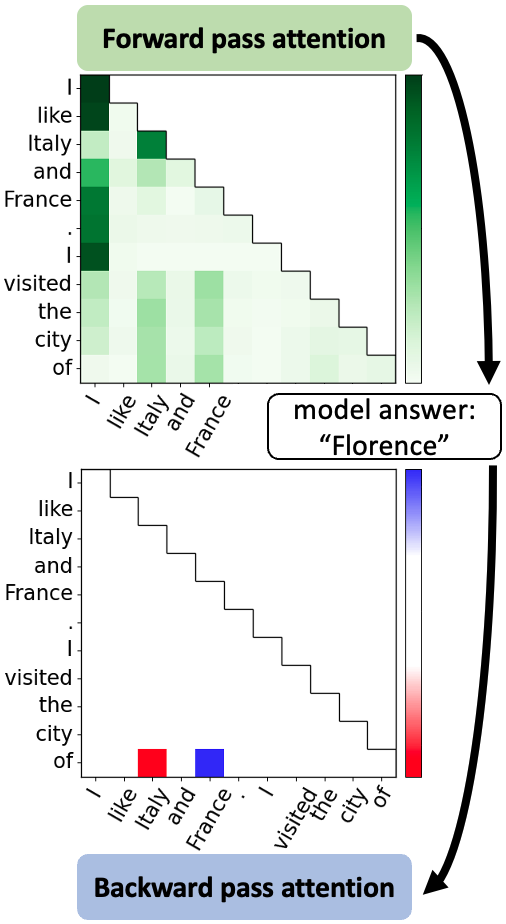}
  \vspace{-.91cm}
  \rule[-.5cm]{4cm}{0cm}
  \caption{In this paper we examine the attention maps obtained from the backward pass, which we named ``Reversed Attention'' (RA). This example present the forward and backward pass of a single attention head of GPT2-xl when prompt with ``I like Italy and France, I visited the city of''. After the model answer ``Florence'', a city in Italy, we apply a backward pass with ``Florence'' as the target for the loss and produce the RA maps. Between all the 1200 attention heads this model has, the presented head has the highest RA map's norm. Compared to the forward attention map, the RA map is more sparse and interpretable. This RA demonstrates how the backpropagation attempts to amplify the information from the token ``Italy'' (red) while reducing the influence of ``France'' (blue).
  }
\label{fig: RA first page}
\end{figure}

\section{Introduction}
The widespread use of automatic gradient technologies such as AutoGrad~\citep{autograd2016modeling, pytorch2019} to obtain the gradients may cause the explicit derivations of these gradients to be overlooked. In this work we explore the equations that govern the backpropagation of the popular Transformer-based~\citep{vaswani2017attention} GPT~\citep{radfordimproving} architecture.

This explicit analysis leads to several surprising discoveries. First, while during the forward pass the attention mechanism explicitly creates triangular attention matrices, from multiplying the query and key vectors, during the backward pass it implicitly creates triangular matrices that determine the gradients of the queries and keys. Due the similarity of these triangular matrices to the forward attention, we refer to them as the Reversed Attention (RA) matrices, \autoref{fig: RA first page}.
Secondly, we study the effect induced by GD and how it tries to increase or decrease attention scores between the queries and keys of each attention head.

Based on these discoveries, we explore the use of RA in LM explainability. While the forward pass attention, which fails in providing a clear explanation of the model's behavior \citep{jain2019attention}, we demonstrate RA's ability to achieve competitive results in concert with methods such as causal mediation \citep{vig2020investigating, meng2022locating} in perturbation analysis.
Furthermore, one can view the RA as a correction term to the attention, given the loss of the backward pass. As an application, we inject the RA scores directly into the forward pass of attention, in order to modify the model's predictions.
This novel method, which we call ``attention patching'', does not involve any parameter updates and offers a new perspective on how interventions can be performed on LMs.

Our main contributions are as follows:
(i) We provide a mathematical walk-through and interpretation of the gradients and Vector-Jacobian Products (VJPs) governing the backpropagation of GPT.
(ii) We identify the attention-softmax derivative as an implicit attention map, which we term Reversed Attention (RA).
(iii) We visualize and qualitatively explore the interpretability of RA.
(iv) We conduct a perturbation test to quantify the explainability of RA.
(v) We demonstrate a novel patching method that uses RA to edit LM's predictions.

\section{Related Work}
All leading deep learning models are trained using variants of Gradient Descent (GD), an implementation of the backpropagation algorithm. 
While much research examines the impact of GD on GPTs' performance, the internal computations of this process often remain a black box \citep{radfordimproving, Radford2019LanguageMA, gururangan2020don}.
Some studies simplify the Transformer architecture to understand GD, for instance, reducing multi-head attention to a single head or linear attention in toy models \citep{tian2023scan, tarzanagh2023transformers, mahankali2023one, dai2023can}. 
Notably, the literature on the backward pass of full multi-head attention is limited.
Our work addresses this gap by examining assumption-free full GPT models, with a specific emphasis on detailing the mathematical computation of gradients.

Previous investigations into weight updates via GD have focused primarily on the MLP layers or the data it was trained on \citep{gueta2023knowledge}. Recently, \citet{katz2024backward} has revealed that gradients can be interpreted as tokens' embeddings.
Specifically, the Vector Jacobian Products (VJPs) that are passed by the residual from the last layer's loss to earlier layers can be seen as an intermediate state toward the model as it adjusts its weights.

{Gradients have been leveraged to localize the source of specific model behavior for a given input  \citep{simonyan2014deep, ancona2018towards} or specific components of the models \citep{barkan2021grad, ma2023llm}. 
In our study, we explore the use of RA as a means to quantify the relative influence of the various components of a model. We compare our method with the current leading technique for such localization, Causal Mediation (CM) \citep{vig2020investigating, meng2022locating, meng2022mass}. CM involves probing the effect of altering components during the forward pass, which necessitates significantly more computation compared to the RA approach.}

Recent works explore the patching technique, where one model's intermediate state is integrated into the forward pass of another \citep{zhang2023towards, elhage2021mathematical, wanginterpretability, todd2023function}.
While they focus on activation patching, we propose a novel approach: directly injecting attention scoring maps of each attention head, without parameter updates.

As far as we can ascertain, we are the first to identify, visualize, and explore the dynamics of the backward pass using RA maps.

\section{Background}
\label{Background}
This section provides the necessary background and establishes the notation used.

Generative Pre-trained Transformer (GPT) is an auto-regressive architecture of multiple transformer blocks connected via a residual stream.
As input, a GPT model receives a sequence of $n$ tokens (a prompt) and predicts a single token.
An embedding matrix at the start of the model embeds the token into vectors $X=[{x^1,\cdots,x^n}]\in \mathbb{R}^{n\times d}$ where $d$ is an embedding dimension.
At the end of the model, the embedded predictions are projected back into tokens using a decoding matrix.
Each transformer block consists of two sub-blocks: multi-head attention (Attn) and a Multi-Layer Perceptron (MLP), interconnected by a residual stream.

The attention mechanism is executed using matrices $W_q, W_k, W_v, W_o \in \mathbb{R}^{d\times d}$, named query, key, value, and output, respectively.
This calculation is performed after splitting the matrices vertically into $h$ non-overlapping parts, called heads.
We denote the attention matrices for the $l$ head in $\hat{W}^l$, hence $\hat{W}^l_q, \hat{W}^l_v, \hat{W}_o^{l, \top} \in \mathbb{R}^{d\times \frac{d}{h}}$, and for example $W_q=[\hat{W}^1_q, \cdots, \hat{W}^h_q]$.
The first three matrices are used to project the input into queries, keys, and values:
\begin{align}
    Q^l=X \hat{W}_q^l, \  K^l=X \hat{W}_k^l, \  V^l=X \hat{W}_v^l \in \mathbb{R}^{d\times \frac{d}{h}}
\end{align}

Together, the queries and keys are used to calculate the forward attention scores:
\begin{align}
    A^l = \text{softmax} \left( \frac{{Q^l K^{l \top} }}{{\sqrt{d/h}} } + M \right) \in \mathbb{R}^{n\times n}
    \\
     M_{ij} = \begin{cases}
    0 \  & \text{if } i \ge j \\
    -\infty  & \text{otherwise}
\end{cases}
\end{align}
Where $M\in \mathbb{R}^{n\times n}$ is a masking matrix that zeroes all scores but the one representing the connection of an earlier token to further ones: .
The output of each head is calculated by multiplying the attention scores with the values and projecting the result back using the output matrix. The attention block output is the sum of the output of all heads: $\text{Attn}(X) = \sum_{l=1}^h \text{head}^l$, where $\text{head}^l = A^l V^l \hat{W}_o^l$ .

The MLP block is a pair of fully connected matrices $FF_1$, $FF_2^\top \in \mathbb{R}^{d\times d_m}$ and an activation function $f$. The output of this block is: $\text{MLP}(X)=f(X FF_1) FF_2^T$. Lastly, the forward pass of the $i$-th transformer block on its input hidden state, $X^i$, is: $X^{i+1} = X^i + \text{Attn}(X^i) + \text{MLP}(\text{Attn}(X^i)+X^i)$.

Note that GPT models also include Layer Norms. For simplicity and due to the relatively low contribution to the gradients and inconsistency when they are placed within different architectures, we omit them from this explanation.

\paragraph{Gradient decent, backward pass and VJPs} 
GD's backward pass is the execution of Backpropagation \citep{le1988theoretical}, the process of applying the chain rule to compute a model's gradients.
A backward pass is initiated after the model executes a forward pass; it computes a loss score $L$, comparing the model's output with a desired target.
This loss score is propagated back through the model's layers as an error signal, in the reverse order of the forward pass.
The error signal can be represented as a vector that is used as an intermediate state of the backward pass, similar to the hidden state in the forward pass.
Given a model's parameter $W$, which is used to compute $z=xW$, where $x \in \mathbb{R}^{d_1}, z\in \mathbb{R}^{d_2}$,
the error signal is the loss with respect to the layer's output, $\delta=\frac{\partial L}{\partial z}\in \mathbb{R}^{d_2}$.
This vector is known as the {Vector-Jacobian Product (VJP)} of $z$.
At the last layer of the model, the VJP is calculated directly by the loss function.
For earlier layers, the VJP is calculated using the backward step, where the VJP of the next layer is used to compute the VJP of those that precede it.
For instance, the output of a sequential layer $l$ is the input of $l+1$, meaning $z^l=x^{l+1}$.
Given those layers are weight matrices, the VJP of the $l$ layer is computed by the following step:
\begin{equation}
\delta^l=\frac{\partial L}{\partial z^l}=\frac{\partial L}{\partial x^{l+1}}=\delta^{l+1} (W^{l+1})^\top
\end{equation}

Finally, the gradient of each weight matrix $W$ is the outer product of the layer's input $x$ and the VJP $\delta$, which updates the weights using a learning rate $\eta \in\mathbb{R}$:
\begin{align}
\label{eq: grad for one}
\frac{\partial L}{\partial W}=\frac{\partial z}{\partial W}\frac{\partial L}{\partial z}=x^\top \times \delta \in{\mathbb{R}^{d_2\times d_1}}
\\
W \leftarrow W - \eta \frac{\partial L}{\partial W}^\top \in{\mathbb{R}^{d_1\times d_2}}
\end{align}
In models such as GPTs, each forward pass includes a sequence of inputs $X={x^1, \cdots, x^n} \in \mathbb{R}^{n\times d_1}$.
In this case, each input has its own VJP, $\delta^i\in \mathbb{R}^{d_2}$, and the full matrix's gradient is the sum of the outer products of each input and its VJP:
\begin{equation}
\label{eq: grad for all}
\frac{\partial L}{\partial W}=\sum_{i=1}^n x^{i \top} \times \delta^i \in{\mathbb{R}^{d_2\times d_1}}
\end{equation}

\section{Attention Layers Gradients}
\label{Attention Layers Gradients}
In this section, we examine the VJPs and gradient matrices for each of the attention layer matrices. The purpose of this examination is to reveal properties of the gradient matrices, as well as to fill a missing gap in the literature. 
In this mathematical walk-through, we examine the submatrices of each attention head, $\hat{W}_q^l, \hat{W}_k^l, \hat{W}_v^l, \hat{W}_o^l$, dropping the head index $l$ for brevity.
The gradients of the full matrices are the concatenation of the heads' gradients (same as the concatenation of weight matrices).

Throughout this section, we denote the forward pass input and output vectors of each submatrix $\hat{W}\in \mathbb{R}^{d\times \frac{d}{h}}$ by $x\in \mathbb{R}^d$ and $z\in \mathbb{R}^{\frac{d}{h}}$, i.e., $z=xW$. The only exception is $W_o$, where the dimensions are swapped, $\hat{W}_o\in \mathbb{R}^{\frac{d}{h} \times d}$. 
$A_{i,j}$ denotes the forward pass attention score from the $i$-th token to the $j$-th token. Lastly, considering the model's input as a sequence of \(n\) tokens, instead of explicitly writing the gradient of every matrix and token, we will focus on determining the VJP for a single token \(j \in \{1, \cdots, n\}\), annotating its input and VJP with \(x^j, \delta^j\). Given the inputs and VJPs, the gradients are the outer product of the two \autoref{eq: grad for all}.

\paragraph{The output projection matrix $\hat{W}_o$}
The VJPs of \(\hat{W}_o\), denoted as \(\delta^j_o\), are obtained directly from the residual stream at the end of the attention block of \(\hat{W}_o\), denoted by Attn(\(x\)) in \autoref{Background}. 
Viewing GD as an application of the chain rule, we consider $\delta_o^j$ as the layer's intermediate editing target.

Only for this matrix, we will demonstrate that when we are given its VJPs, we can infer its gradient and the effect of GD updating. This explanation holds for all further matrices, too.
The gradient updates introduced by only the $j$-th token (left equation) or by all $n$ tokens (right equation) are:
\begin{align}
\label{eq: Wo update}
\hat{W}_o \leftarrow \hat{W}_o - \eta \delta^j_o \times x_o^{j \top}
\\
\hat{W}_o \leftarrow \hat{W}_o - \eta  \sum_{i=1}^n \delta^i_o \times x_o^{i \top},
\end{align}
where $\eta$ is the update's learning rate. In general, if we simplify the full update and consider only the change introduced by a single token (left equation) to examine a future forward pass with the same input $x_o^j$, then the original layer's output $z^j_o = x_o^j \hat{W}_o $, is shifted by the direction of $\delta_o^j$ with the magnitude of $-\eta \lVert x^j_o \rVert_2^2 \in \mathbb{R}$:
\begin{align}
\label{eq: Wo update2}
x_o^j \hat{W}_{o GD} = x_o^j(\hat{W}_o - \eta \delta^j_o \times x_o^{j \top}) =
\\
z^j_o - \eta \lVert x^j_o \rVert_2^2 \delta^j_o
\end{align}

\paragraph{The value projection matrix $\hat{W}_v$}
In the forward pass, the value vector associated with the $j$-th token is given by $v^j = x_v^j \hat{W}_v \in \mathbb{R}^\frac{d}{h}$. 
To calculate the VJP for the $j$ token, we consider, due to causality, the subsequent tokens $l\geq j$. For every such token, the backward pass propagates its own VJP from $\hat{W}_o$ by computing the error signal from the $l$-th token to token $j$:
\begin{equation}
\label{eq: Wo to Wv}
e^l=\delta_o^l \hat{W}_o^\top \in \mathbb{R}^{\frac{d}{h}}\,,
\end{equation}
where $\delta_o^l$ is the VJP of $\hat{W}_o$ computed for the $l$-th forward pass (the $l$-th token) of the autoregressive process. 
The VJP of $\delta_v^j$ is a weighted sum of the error signal and the forward attention scores $A$:
\begin{equation}
\label{eq: Wv vjp}
\delta_v^j =  \sum_{l=j}^{n} A_{l,j}e^l  \in \mathbb{R}^{\frac{d}{h}}
\end{equation}

\paragraph{Softmax derivative}
Although the softmax function is not a weight parameter of the model, comprehending its role in the backward pass is crucial for understanding the editing dynamics.
In the forward pass, the attention scores for the $j$-th forward pass, $A_j\in \mathbb{R}^{1 \times n}$ (the $j$-th row of the forward attention) are generated by applying the softmax function to the product of the query and key vectors, with a scaling factor of $\sqrt{\frac{d}{h}}$.
These scores are then multiplied by the attention values, $V\in \mathbb{R}^{n \times \frac{d}{h}}$.
The backward pass performs a reverse operation: each token's error signal from $\hat{W}_o$ is multiplied by the attention values and scaled as follows:  
\begin{align}
\label{eq: softmax derevative}
\tilde{e}^j &= \delta_o^j \hat{W}_o^\top V^\top = e^j V \in \mathbb{R}^{n}
\\
r^j &= A_j \odot (\tilde{e}^j - \tilde{e}^j A_j \cdot \mathbf{1}^n) \sqrt{\frac{h}{d}} \in \mathbb{R}^{n},
\end{align}
where $\odot$ is the element-wise product of two vectors (Hadamard product) and $\tilde{e}^j A_j \cdot \mathbf{1}^n$ is the scalar $\tilde{e}^j A_j$ assigned to an $\mathbb{R}^{n}$ vector.

\textbf{Rewriting as a batch of all tokens:}
If we rewrite \autoref{eq: softmax derevative} for all tokens together, by concatenating their VJP into a matrix $\Delta=[\delta_o^1, \cdots, \delta_o^n]\in \mathbb{R}^{n \times d}$ we get:
\begin{align}
\label{eq: reversed attention}
\tilde{E} &= \Delta \hat{W}_o^\top V^\top \in \mathbb{R}^{n \times n} \\
R &= A \odot \left( \tilde{E}^\top - \textit{diag}(A \tilde{E}^\top)\right)^\top \sqrt{\frac{h}{d}} \in \mathbb{R}^{n \times n},
\end{align}

hence $R_j=r^j, \tilde{E}_j=\tilde{e}^j$, where $_j$ is the $j$-th row of each matrix. 
We further use the derivative $R$ to compute the subsequent gradients of $\hat{W}_q, \hat{W}_k$. In \autoref{Reversed Attention} we investigate what $R$ represents.

\paragraph{The query projection matrix $\hat{W}_q$}
During the $j$-th forward pass, the $j$-th query is generated by computing $q^j = x_q^j \hat{W}_q \in \mathbb{R}^\frac{d}{h}$. The query is then multiplied by all key vectors, $K=[k^1,\cdots,k^n] \in \mathbb{R}^{n \times \frac{d}{h}}$ (and masking the ones that followed it), to obtain the raw attention scores (logits). Therefore, the backward pass calculates the VJP of the query by calculating:
\begin{equation}
\label{eq: Wq VJP}
\delta_q^j = r^j K = R_j K \in 
\mathbb{R}^{\frac{d}{h}}
\end{equation}

\begin{figure*}[h!]
    \begin{tabular}{ccc}
\includegraphics[width=.35\linewidth]{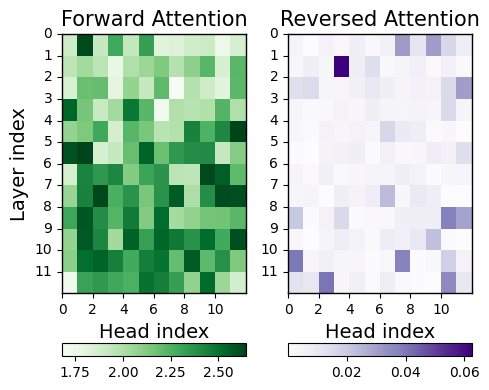} & 
        \includegraphics[width=.31\linewidth]{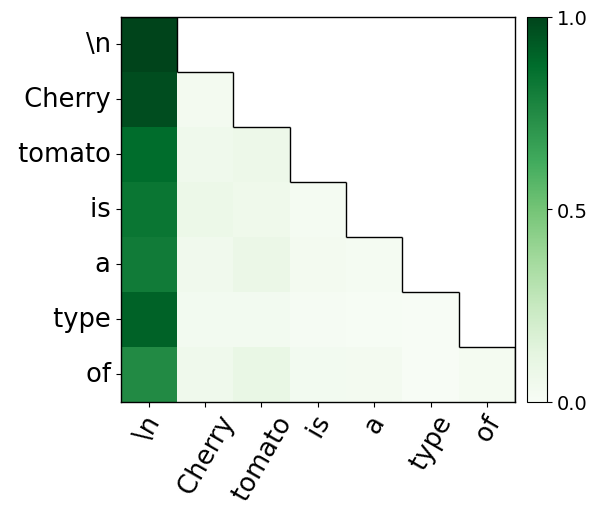} & 
        \includegraphics[width=.33\linewidth]{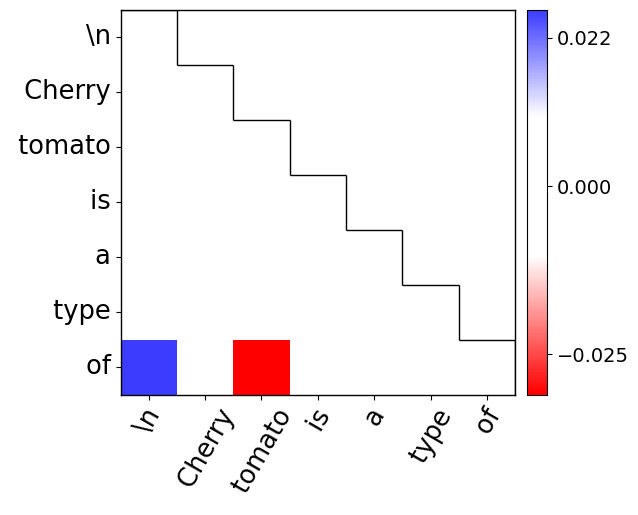} \\ 
        (a) & (b) & (c)
    \end{tabular}
    \caption{(a) The norms of the attention maps per head and per layer. (b) Forward and (c) Reversed Attention of the same head from GPT2-small (layer 11, head index 2). This is the attention head with the second highest Reversed Attention norm and we can see it focused on editing the query of ``of'' (row) and the key of ``tomato'' (column).}
    \label{fig:FAandRA}
    \vspace{-.25cm}
\end{figure*}

\begin{figure}[h!]
\centering
\begin{minipage}[b]{0.41\textwidth}
    \centering
    \begin{overpic}[width=\linewidth]{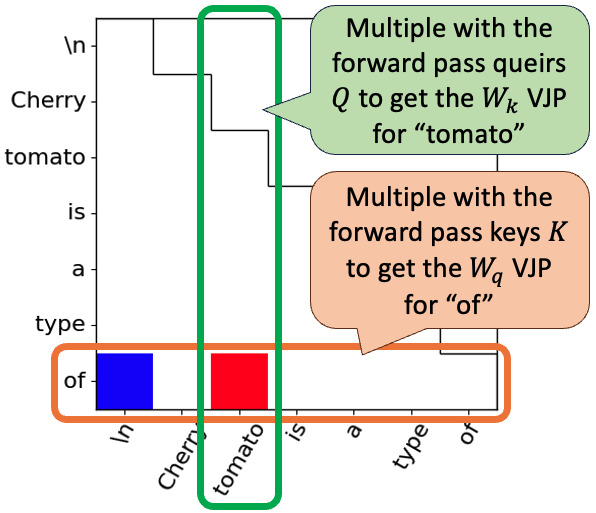}
            \put(-7,70){\textbf{(a)}}
              \end{overpic}
\end{minipage}%
\hfill
\begin{minipage}[b]{0.45\textwidth}
    \centering
    \begin{overpic}[width=\linewidth]{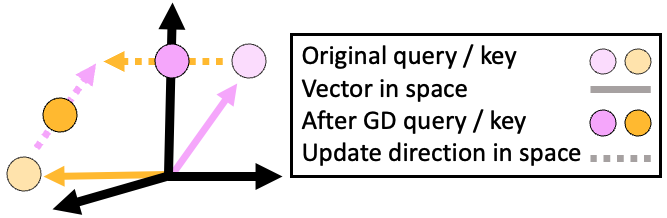}
            \put(0,25){\textbf{(b)}}
        \end{overpic}
    \captionof{figure}{RA model editing dynamics. (a) The query matrix \(\hat{W}_q\) will be updated with a VJP directed towards the forward pass key of ``tomato'', while the key matrix \(\hat{W}_k\) will be updated with a VJP directed towards the query from the token ``of''. (b) The latent space of the queries and keys. The circles represent a forward pass query and a key. If their Reversed Attention score is a relatively low negative number, the directions they are moving towards after GD are actually towards one another.}
    \label{fig:raedit}
\end{minipage}
\end{figure}

\paragraph{The key projection matrix $\hat{W}_k$}
The gradients of \(\hat{W}_k\) are computed similarly to those of \(\hat{W}_q\), except that for the \(j\)-th forward pass, we utilize the queries \(Q=[q^1,\cdots,q^n] \in \mathbb{R}^{n \times \frac{d}{h}}\) from each subsequent forward pass after \(j\).
The VJP and gradients of $\hat{W}_k$ are then given by:
\begin{equation}
\label{eq: Wk VJP}
\delta_k^j = \sum_{l=j}^{n} r^l_j  x_q^l \hat{W}_q = \sum_{l=j}^{n} r^l_j q^l =  R^\top_j Q \in 
\mathbb{R}^{\frac{d}{h}} ,
\end{equation}
where $R^\top_j$ is the $j$-th column in $R$, which is the error signal from the $l$-forward pass to the $j$-th one.

\section{Reversed Attention}
\label{Reversed Attention}
In \autoref{eq: reversed attention} we defined $R$ as the softmax derivative. $R$ shares many properties with $A$, the forward attention (FA):
\begin{itemize}
\item \(R\) is computed from the multiplication of the forward pass values \(V\) and the error signal \(\Delta \hat{W}_o\). We denote this intermediate result as \(\tilde{E}\). The \(i,j\) entry in \(\tilde{E}\) is a score between the \(i\)-th error signal and the \(j\)-th forward pass value. This resembles the calculation of raw attention scores (logits, before softmax) where we multiply queries and keys to obtain scores between every pair of tokens.
\item After computing \(\tilde{E}\), we derive \(R\) from it by row-wise normalization, which involves scaling with the FA \(A\), see \autoref{eq: softmax derevative}. In the forward pass, row-wise normalization is accomplished by applying the softmax function.
\item Since the normalization from \(\tilde{E}\) to \(R\) includes element-wise multiplication with the FA \(A\), which is a lower triangular matrix, \(R\) is also a lower triangular matrix.
\item Just as the FA is used to multiply the attention values $V$, we use \(R\) to multiply the forward pass' queries $Q$ and keys $K$ to obtain the VJPs of one another.
\end{itemize}
These shared properties between the FA and the softmax derivative suggest that the softmax derivative serves as an implicit attention matrix. We called \(R\) the ``Reversed Attention'' (RA).
In the following sections, we delve into some properties of the RA and explore its potential uses in explaining and controlling GPTs. 

\begin{table*}[h!]
  \centering
  \begin{small}
  \begin{tabular}{@{}l@{~~}l@{~}c@{~}c@{~~~}c@{~~~}c@{~~~}c@{~}c@{~~~}c@{~~~}c@{~~~}}
    \toprule
    & &  \multicolumn{4}{c}{1-shots ICL} & \multicolumn{4}{c}{5-shots ICL}                    \\
     \cmidrule(lr){3-6}
     \cmidrule(lr){7-10}
     
    Task          & Example & Random & CM & FA & RA & Random & CM &  FA & RA \\
    \midrule
antonym &  Q: output\textbackslash nA: [input] & 0.02 & \textbf{0.15} & 0.02 & 0.07 & 0.09 & 0.27 & 0.05 & \textbf{0.32} \\ 
alphabetically-first &  Q: finch, tender, peacock\textbackslash nA: [finch]  & 0.12 & \textbf{0.22} & 0.08 & 0.2 & 0.15 & 0.19 & 0.08 & \textbf{0.23} \\ 
choose-middle-of-3 &  Q: dress, paintbrush, vase\textbackslash nA: [paintbrush]  & 0.1 & \textbf{0.35} & 0.08 & 0.19 & 0.17 & 0.21 & 0.1 & \textbf{0.3} \\ 
country-capital &  Q: Sierra Leone\textbackslash nA: [Freetown]  & 0.07 & 0.22 & 0.09 & \textbf{0.29} & 0.19 & 0.4 & 0.07 & \textbf{0.42} \\ 
next-item &  Q: XV\textbackslash nA: [XVI]  & 0.06 & \textbf{0.25} & 0.04 & 0.14 & 0.17 & 0.31 & 0.07 & \textbf{0.34} \\ 
person-sport &  Q: Scottie Pippen\textbackslash nA: [basketball]  & 0.22 & 0.35 & 0.1 & \textbf{0.37} & 0.21 & 0.34 & 0.09 & \textbf{0.44} \\ 

    \bottomrule
  \end{tabular}
  \end{small}
  \caption{GPT2-xl perturbation testing on ICL tasks, measuring the AUC for Reversed Attention (RA), Forward Attention (FA), Casual Mediation (CM) as well as random ordering of the attention heads. For each example, ``[]'' is the token we expect the model to return.}
  \label{tab: perturbation main icl}
\medskip
  \centering
  \begin{small}
  \begin{tabular}{@{}l@{~}l@{~}c@{~~}c@{~~}c@{~~}c@{~~}c@{~~}c@{~~}c@{~~}c@{~~}@{}}
    \toprule
     & &  \multicolumn{4}{c}{GPT2-xl} & \multicolumn{4}{c}{Llama2-7B}                    \\
      \cmidrule(lr){3-6}
      \cmidrule(lr){7-10}
     
    Task          & Example & Rand & CM & FA & RA & Rand & CM & FA & RA\\
    \midrule

country-capital &  The capital city of Sierra Leone is [Freetown]  & 0.02 & \textbf{0.61} & 0.06 & 0.07 & 0.31 & 0.05 & 0.07 & \textbf{0.43} \\ 
person-plays-pro-sport &  Scottie Pippen plays the sport of [basketball]  & 0.32 & \textbf{0.59} & 0.09 & 0.57 & 0.21 & 0.39 & 0.04 & \textbf{0.48} \\ 
product-by-company &  Blogger was created by [Google]  & 0.19 & \textbf{0.46} & 0.08 & 0.31 & 0.15 & \textbf{0.51} & 0.06 & 0.31 \\ 
    \bottomrule
  \end{tabular}
  \end{small}
  \caption{GPT2-xl and Llama2-7B perturbation tests on natural questions, measuring the AUC for Reversed Attention (RA), Forward Attention (FA), Casual Mediation (CM) and random ordering.}
  \label{tab: perturbation main natural}
\end{table*}

\subsection{A qualitative examination of Reversed Attention}
\label{A qualitative examination}
We observe the behavior of RA through a single example, using GPT2-small \citep{Radford2019LanguageMA}. 
We prompt the model with the sentence ``\textbackslash  Cherry tomato is a type of', and using ``tomato'' as an editing target. 
Since this prompt already contains the token answer, the behavior of the model can be readily understood.

\autoref{fig:FAandRA}(a) depicts, for each attention head and each layer, the norm of the FA and the RA. While RA displays a very sparse pattern, FA (which is normalized by the softmax) displays a larger number of heads with high values. There does not seem to be a correlation between the two maps, and this is further supported by the perturbation experiments in \autoref{Reversed Attention identifies attention heads' importance}.

Our mathematical analysis in \autoref{eq: grad for one}, \autoref{Attention Layers Gradients} implies that close to zero scores in the RA will produce close to zero VJP vectors and gradients, hence focusing on the attention heads with the highest RA norm is informative enough to update the GD steps.

Next, we consider one of the FA maps with the highest RA norm, which is provided in \autoref{fig:FAandRA}(b). The key of the first token, ``\textbackslash n'', receives the highest attention scores, which is a well-known phenomenon \citep{xiao2023efficient}.

On the other hand, the RA in \autoref{fig:FAandRA}(c) is sparse and shows that the row corresponding to the last token ``of'' stands out as much more dominant than the others.
We recall that the VJPs of the query matrix $\hat{W}_q$ are the multiplication of the RA with the forward pass keys $K$. Similarly, the VJPs of the keys matrix $\hat{W}_k$ are the multiplication of the RA with the forward pass queries $Q$. Since GD updates are performed with a negative learning rate, positive scores in the RA shift the model's weights towards the queries/keys that produced them (and vice versa).
Hence, the VJP of the query for ``of'' is mostly directed towards the key of ``tomato'' while directed away from the key that belongs to ``\textbackslash n''. 
Similarly, the main update is to the key of ``tomato'', which attempts to shift the model's weights towards the query for ``of''.
This dynamic is illustrated in \autoref{fig:raedit}.
This example does not necessarily elucidate the function served by this attention head, but it demonstrates how GD attempts to repurpose this head to recall information from the token ``tomato''.

This example demonstrates the potential of Reversed Attention to provide insights into the model's behavior and editing dynamics.
Additional examples are provided in \autoref{Additional Reversed Attention examples}.

\subsection{Reversed Attention and the importance of each attention head}
\label{Reversed Attention identifies attention heads' importance}
In \autoref{A qualitative examination}, we demonstrate how we can interpret the effect of RA on the editing dynamics of the model.
This explanation is based on the assumption that high RA scores correspond to important components (model parameters) in the forward computation graphs.

To verify this assumption, we conduct a perturbation test. This test compares different orders (rankings) of the attention heads, each produced by a method that aims to determine the relative importance between heads. This experiment begins by zeroing out (masking) all heads, resulting in poor performance. Gradually we unmask the heads according to a given order. The performances can be quantified by the Area Under the Curve (AUC) of the graph that displays the model's accuracy as a function of the percentage of unmasked heads.

For RA, we rank the heads according to the RA norms (from highest to lowest). 
Additionally, we use a training set (omitted examples) to perform separate backward passes and average the norms of each head.
The main method we compare to is Causal Mediation (CM), due to its extensive usage in interpreting LM \citep{meng2022locating, mueller2024quest}.
This method examines causal indirect effects by patches the forward pass attention heads' outputs and examines the disparity between the altered and the original model's outputs.
In addition, we include the forward pass attention (FA). Similar to RA, FA ranks attention heads according to the the norm of the attention maps. While previous works already established that FA is not sufficient in providing similar models' explanation \citep{serrano2019attention, jain2019attention}, we include FA for completeness.

Our tests are conducted on 21 tasks by \citet{hernandez2024linearity, todd2023function}, each consisting of pairs of short sentences that follow some relation. For instance, one task involves pairs of countries and their capital cities. In 6 of these tasks, the model is prompted with natural language templates such as ``\textit{The capital city of <country name> is}''. In the other 15 tasks, in-context-learning (ICL) templates are employed. In this scenario, instead of explicitly stating the relation between the pair (e.g., ``the capital city is''), the model must infer it from $n$-shots of labeled pairs, such as ``{\textbackslash n Q: Spain A: Madrid \textbackslash n Q: Italy A: Rome \textbackslash n <country name> A:''.

The results are shown for the ICL tasks and the natural questions, respectively, in \autoref{tab: perturbation main icl} for GPT2-xl and \autoref{tab: perturbation main natural} for both GPT2-xl and Llama2-7B \citep{touvron2023llama}. The full implementation details and results, including ones for GPT-j \citep{gpt-j} and OPT \citep{zhang2022opt}, can be found in \autoref{Perturbation test}.

As can be seen, RA is competitive with CM when the examined LM can successfully address the task.
In the natural language tasks, we observed that RA outperforms CM with larger models that originally had high accuracy. In the ICL tasks, we found that CM achieves good results with a very low number of shots when the LM fails to provide the correct answer. However, when we prompt these failed tasks with a few shots, we see that RA achieves better results.
Overally, even when RA falls behind CM, it still achieves non-trivial results. Hence, we find RA to reflect the importance of each attention head in producing a given prediction.
We note that RA is much faster, as it only requires a forward and a backward pass for each example, while CM requires a forward pass for each head.

Despite not setting new state-of-the-art results, the comparison with CM and FA highlights that RA is an interpretable method.

\begin{figure*}
  \vspace{-1cm}
  \centering
  \rule[-.5cm]{0cm}{4cm}
  \includegraphics[width=1.95\columnwidth]{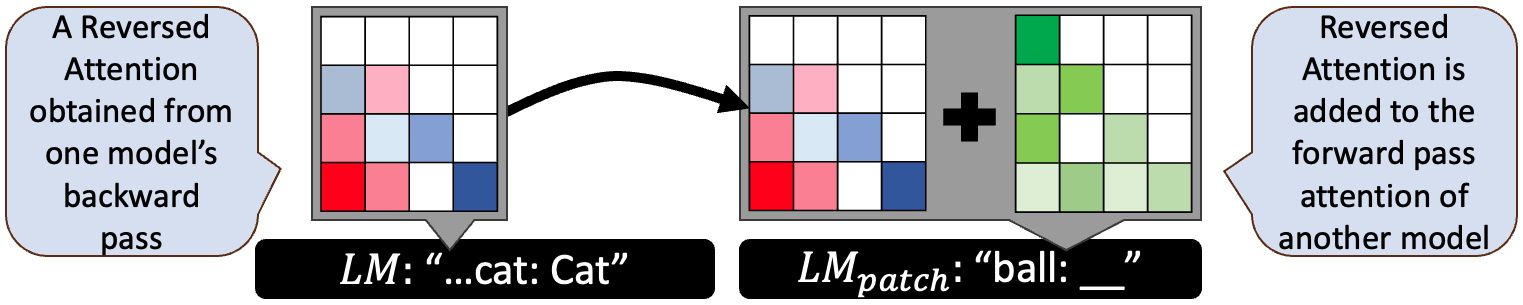}
  \vspace{-.91cm}
  \rule[-.5cm]{4cm}{0cm}
  \caption{Attention patching using Reversed Attention (RA): first we collect the RA maps of the model without applying any model editing (without changing its weights). Later, for each attention head, we add its corresponding RA map to the forward pass attention.
  }
\label{fig: RA editing}
\end{figure*}

\subsection{Attention patching using Reversed Attention}
\label{Attention patching using Reversed Attention}
Modifying the forward attention maps of LM to improve them, while doing so in an explainable manner, is one of the goals of interpretability research. However, since LMs can have thousands of attention heads, each depicting different relationships, a solution that directly edits the attention maps has not been studied extensively. In this section, we demonstrate how RA can achieve this goal.

\begin{table}[h]
  \caption{GPT2-xl and OPT-1.3B accuracy on ICL tasks of the original models and with forward attention (FA) and Reversed Attention (RA) patching. N = the number of ICL samples.}
  \label{tab: patching main}
  \centering
  \begin{small}
  \begin{tabular}{@{}l@{~~}l@{~~}c@{~~}c@{~~}c@{~~}c@{~~}c@{~~}c@{~~}c@{}}
    \toprule
        & &  \multicolumn{3}{c}{GPT2-xl} & \multicolumn{3}{c}{OPT-1.3B} 
        \\
     \cmidrule(lr){3-5}
     \cmidrule(lr){6-8}

     Task & N & original & FA & RA &
                original & FA & RA \\ 
\midrule 
        \multirow{4}{3.5em}{antonym} & 0 & 0.00 & 0.01 & \textbf{0.08} & 0.02 & 0.01 & \textbf{0.24} \\
       & 1 & 0.18 & 0.43 & \textbf{0.56} & 0.20 & 0.26 & \textbf{0.57} \\
       & 5 & 0.53 & 0.57 & \textbf{0.62} & 0.42 & 0.44 & \textbf{0.59} \\
       & 10 & 0.57 & 0.57 & \textbf{0.62} & 0.42 & 0.43 & \textbf{0.54} \\
\midrule 
        \multirow{4}{3.5em}{capitalize} & 0 & 0.00 & 0.00 & \textbf{0.94} & 0.01 & 0.00 & \textbf{0.78} \\
       & 1 & 0.44 & 0.50 & \textbf{1.00} & 0.01 & 0.01 & \textbf{0.90} \\
       & 5 & 0.98 & \textbf{1.00} & \textbf{1.00} & \textbf{1.00} & 0.99 & \textbf{1.00} \\
       & 10 & 0.99 & \textbf{1.00} & \textbf{1.00} & \textbf{1.00} & 0.99 & \textbf{1.00} \\
\midrule 
        \multirow{4}{3.5em}{choose-middle-of-3} & 0 & 0.46 & 0.30 & \textbf{1.00} & 0.11 & 0.03 & \textbf{1.00} \\
       & 1 & 0.81 & 0.76 & \textbf{1.00} & 0.57 & 0.54 & \textbf{0.76} \\
       & 5 & 0.92 & 0.68 & \textbf{1.00} & 0.95 & 0.84 & \textbf{1.00} \\
       & 10 & 0.97 & 0.00 & \textbf{1.00} & 0.86 & 0.65 & \textbf{1.00} \\
\midrule 
        \multirow{4}{3.5em}{next-item} & 0 & 0.03 & 0.00 & \textbf{0.16} & 0.09 & 0.00 & \textbf{0.53} \\
       & 1 & 0.28 & 0.66 & \textbf{0.72} & 0.50 & 0.47 & \textbf{0.84} \\
       & 5 & 0.69 & 0.84 & \textbf{0.88} & 0.69 & 0.66 & \textbf{0.88} \\
       & 10 & 0.88 & 0.88 & \textbf{0.91} & 0.75 & 0.81 & \textbf{0.84} \\

    \bottomrule
  \end{tabular}
  \end{small}
\end{table}

Recent efforts in interpretability research have explored activation patching, demonstrating how injecting different hidden states from one model into another affects its performance. In this section, we explore both the idea of directly modifying attention maps and activation patching through a novel method termed ``attention patching''. This approach is based on the observation that RA produces attention maps that can be seen as the desired relationships the model attempts to maintain in order to perform a given task.

Given a GPT model and a predefined set of training and test examples, all with the same length and format, attention patching performs the following steps: 
(i) calculate the RA of each training example (applying forward and backward passes with each example but without modifying the model's parameters).
(ii) average the RA scores for each attention head.
(iii) For each test example and for each attention head, 
we add (inject) the RA map to the forward pass attention map, using a learning rate as a scaling factor, 
\autoref{fig: RA editing} illustrates this process.
Additionally, to establish a baseline, we applied the same process with the forward attention, averaging and injecting the attention maps collected from forward passes.

The requirement for examples to have the same length and format simplifies the injection process, ensuring that all attention score matrices are of consistent size.
This makes tasks such as ICL and short trivia-like questions ideal candidates for this method, as they are templated and easily framed by their length.
Additionally, the ability of LMs to answer ICL tasks is usually associated with the attention layer \citep{dong2022survey}, which serves as another reason to examine attention patching in this context.

We evaluate attention patching on the datasets of \citep{todd2023function, hernandez2024linearity}, comparing its performance to ICL prompting.
% ... 
The results are displayed in \autoref{tab: patching main}, with additional details provided in \autoref{Attention patching}. 
Our findings indicate that attention patching achieves similar results to ICL prompting and outperforms the average of the forward pass attention scores, {which does not consistently improve the original model performances.}
Note that similarly to other patching methods, this test is not meant to demonstrate the robustness of the approach, as the current implementation is frame-specific and full GD outperforms it. 
Instead, it serves to validate that Reversed Attention indeed reflects the model's desired attention.

\section{Conclusions}
The self-attention component of transformers is perhaps the most distinctive part of this architecture. Its role when performing inference has been extensively studied and shown to provide insights into the inner workings of transformers. Here, we explore a dual entity we call Reversed Attention (RA), which plays a role when transformers learn. We present qualitative samples of the way learning occurs and show that RA can help identify the most influential heads at inference time. Finally, we show how plugging an average RA value can direct the model toward performing a specific task.

The focus of this work was to introduce RA and prove that it is interpretable rather than merely an artifact of the backward pass. Beyond enriching our understanding of how LMs work, we hope our conclusions will serve future research on dynamically editing attention layers as well as explaining how information is stored in them.

\clearpage

{\section{Limitations}}
\label{Limitations}
In this work, we provide a detailed mathematical exposition of the derivative of the attention mechanism in GPTs.
To keep this explanation clear, we focus on decoder-only models with Multi-Head Attention mechanisms and without additional components such as RoPE \cite{su2024roformer} or sparse attention \cite{brown2020language}.
LLMs come in various sizes and configurations, and a general mathematical explanation that fits all is not feasible.
Our choice to demonstrate RA using GPT2 and OPT aligns with previous work that examines the interpretability of Transformers through the lens of these models \cite{geva2022transformer, meng2022locating, voita2023neurons, katz2023visit}. 
Our use of Llama2 and GPT-j, although they use RoPE, comes to show how our use of RA can be applied to a wide variety of LM.

The 21 tasks we sourced from \citep{todd2023function, hernandez2024linearity} consist of relatively simple and limited tasks. Therefore, our results in \autoref{Reversed Attention identifies attention heads' importance} and \autoref{Attention patching using Reversed Attention} serve as a proof of concept rather than as a definitive assessment of the robustness of RA in its ability to identify critical components of models or as editing it using patching.

Causal Mediation (CM) can come in variety of implementations, each patch the activation differently. In \ref{Reversed Attention identifies attention heads' importance} we examined a basic implementation of CM. In \autoref{Perturbation test} we include additional implementation provided by \citep{todd2023function} to make our comparison to RA comprehensive.

The perturbation experiments compare RA with a small number of alternative methods. We acknowledge that other methods might achieve similar or even better results, particularly those based on gradients. The purpose of these experiments is not to discover a new component-localization method but rather to provide proof that the RA maps correspond to existing LM explainability methods. Therefore, we focused on comparing RA to CM, which is a widely used method in similar works.

\smallskip
\medskip
\section*{Ethics Statement}
This paper aims to advance our understanding of how language models learn and the dynamics behind the backward pass. Future work might leverage our findings to edit or train language models effectively. Our hope is that these efforts will contribute to the creation of better and more aligned models, rather than facilitating the production of harmful content or exploiting the knowledge stored within these models.

\section*{Acknowledgements}
This work was supported by the Tel Aviv University Center for AI and Data Science (TAD).

\bibliography{anthology,custom_ref}
\bibliographystyle{acl_natbib}

\appendix

\clearpage

\section{Additional Reversed Attention Examples}
\label{Additional Reversed Attention examples}
In this section, we extend the qualitative examination of the Reversed Attention (RA) discussed in \autoref{A qualitative examination}.
We provide examples for GPT2-small and GPT2-xl.
RA offers simple-to-read insights into how the attention matrices are modified by GD, revealing which attention queries $Q$ and keys $K$ are adjusted to bring specific tokens closer in the embedding space.
However, while this explanation sheds light on the changes made by GD, it does not necessarily elucidate the function served by each attention head.
In this section, we aim to establish how we interpret (read) the patterns observed in RA maps.
Further investigation into the functional roles of individual attention heads is left for future work.

\begin{figure}
  \centering
  \includegraphics[width=0.99\columnwidth]{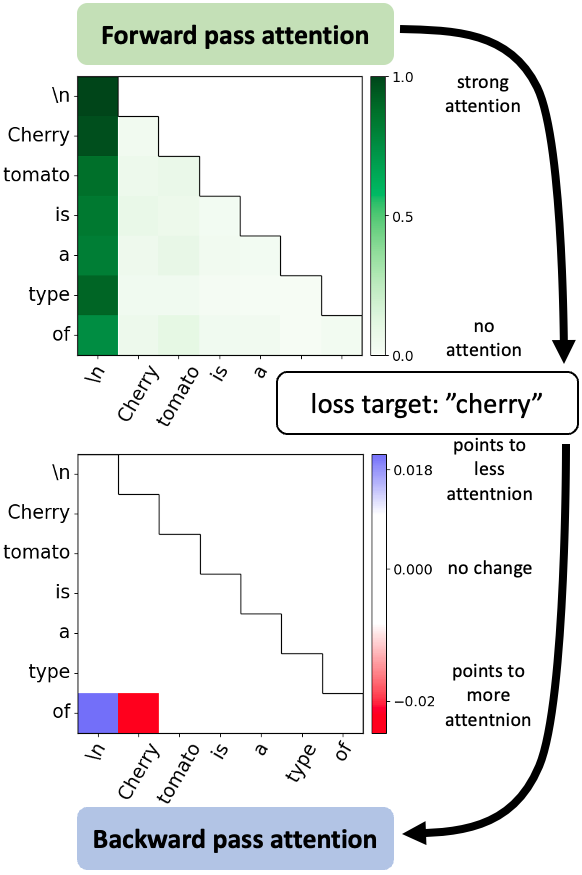}
  \caption{The forward and Reversed Attention (RA) maps of an attention head from GPT2-small (layer 11, head index 2), given the editing target ``cherry'' with the prompt ``Cherry tomato is a type of''. The pattern presented by the RA map attempts to increase the forward pass attention between the query belonging to ``of'' and the key of ``Cherry'', encouraging the model to answer ``cherry''.
  }
\label{fig: forward and reversed attn - cherry tomato 2}
\end{figure}

\paragraph{GPT2-small}
The qualitative example in \autoref{A qualitative examination} presents GPT2-small RA while using the prompt ``The cherry tomato is a type of'' with the editing target ``tomato''.
The pattern given by one of the highest RA norms shows that GD focuses on editing the query belonging to the token ``of'' and the key of the token ``tomato''.
The following example examines what happens if we change the editing target to ``cherry'', which, like ``tomato'', could refer to the prompt itself.
Similarly to the previous example, this head is still one of the heads with highest RA norms, but unlike the previous example, GD focuses on editing the key of the token ``cherry'', as presented in \autoref{fig: forward and reversed attn - cherry tomato 2}.
This pattern, identifies the token most similar to the editing target and comes to illustrate RA's ability to localize relevant information in the input tokens regarding the editing target (``tomato'' or ``cherry''), through the attention heads.

\paragraph{GPT2-xl}
We prompt the model with the following sentence: ``I like Italy and France. I visited the city of'', expecting the model to complete it with a city from either Italy or France.
In the case of GPT2-xl, it returns ``Florence''.
We extract RA when given ``Paris'' as the editing target and examine the top by norm RA heads, head 8 at layer 30.
The forward attention of this head assigns relatively high attention to ``Italy'' and ``France''.

Note that this is the same example from \autoref{fig: RA first page}. Thus, these examples also examined how different targets affect the produced RA maps.

Looking at the RA, the last row, corresponding to the last (``of'') token's \( W_q \) VJP, we see it assigns a positive score to ``Italy'' and a negative score to ``France''.
According to \autoref{Attention Layers Gradients}, \autoref{A qualitative examination}, this pattern suggests that GD tries to bridge the query of the last token with the key of ``France'', while doing the opposite with ``Italy''.
Hence, if we isolate the possible outcomes of editing other heads and only editing this one, its output would be more correlated with the attention value from the token ``Paris''.
We assume this behavior arises from GD's attempt to enhance connection to information that is more relevant to the editing target, ``Paris''.

We repeat the same experiment with the editing target ``Rome''. This time, the RA of the same head assigns a negative score to the token that follows ``Paris'' and a positive score to the token ``Rome''.
Using the same analogy, given the target that is more related to Italy, the model tries to amplify the connection between the last token's query and the key of ``Italy''.
This suggests that the RA method dynamically adjusts attention scores to strengthen the associations relevant to the given editing target.

\begin{figure*}
  \centering
  \includegraphics[width=\linewidth]{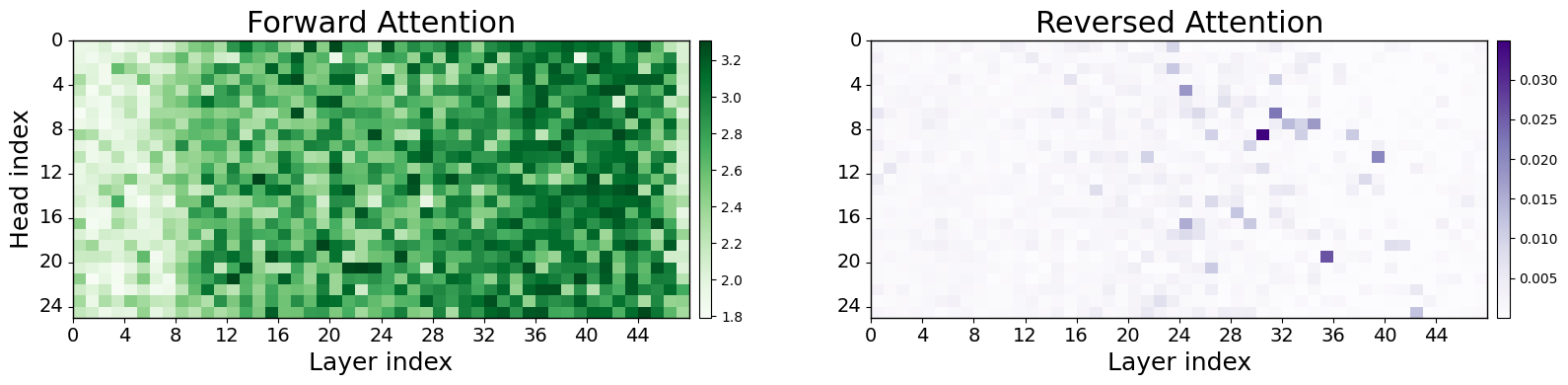}
  \caption{The forward and Reversed Attention maps of GPT2-xl, given the editing target ``Paris'' and the prompt ``I like Italy and France. I visited the city of''.
  }
\label{fig: gpt2xl norms i like}
\end{figure*}

\begin{figure*}
    \centering
    \begin{subfigure}[h]{0.40\textwidth}
        \centering
        \includegraphics[width=\linewidth]{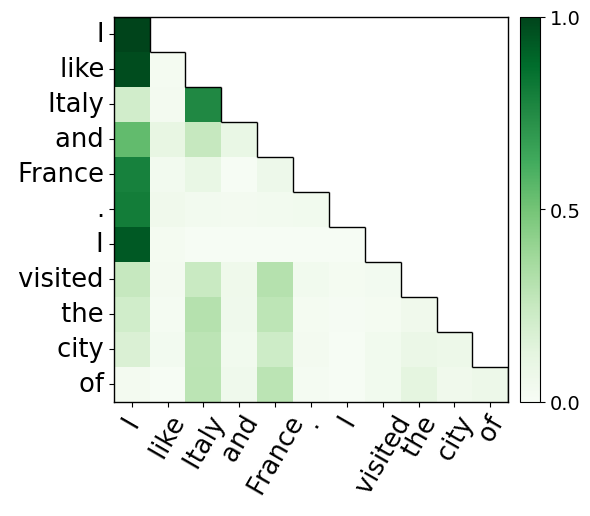}
        \caption{Forward - Florence, Paris and Rome}
        \label{subfig: Forward - Paris}
    \end{subfigure}
    \hspace{10cm}
    \begin{subfigure}[h]{0.40\textwidth}
        \centering
        \includegraphics[width=\linewidth]{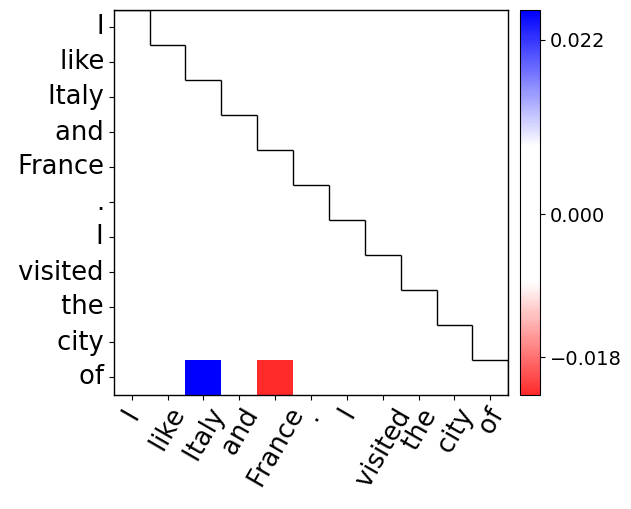} 
        \caption{Reversed - Paris}
        \label{subfig: Reversed - Paris}
    \end{subfigure}
    \begin{subfigure}[h]{0.40\textwidth}
        \centering
        \includegraphics[width=\linewidth]{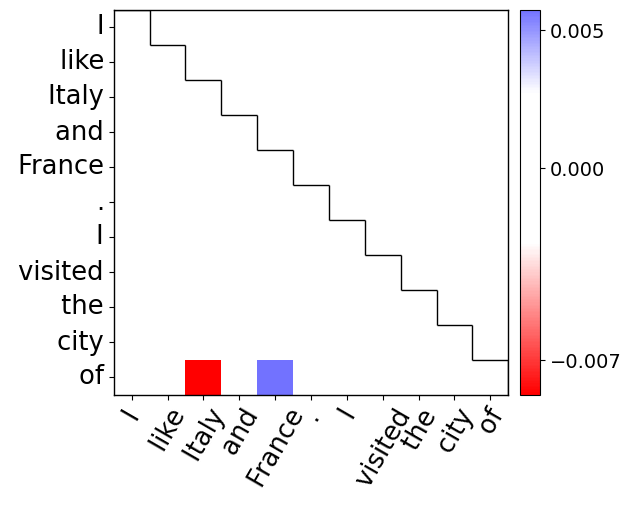} 
        \caption{Reversed - Rome}
        \label{Reversed - Rome}
    \end{subfigure}
    \caption{Attention maps for the prompt ``I like Italy and France. I visited the city of'' (head 8, layer 30). While the forward attention is the same for both cases (a), the changes of the editing target produce two different Reversed Attention maps. Each reversed map shows that GD tries to change the attention block's matrices in order to amplify the attention scores of the tokens that are more correlated to the editing target: by amplifying ``France'' for the target ``Paris'' (b), or by enhancing ``Italy'' when the target is ``Rome'' (c).}
    \label{fig: gpt2xl forward and reversed attn i like}
\end{figure*}

\paragraph{OPT} We use OPT-350m with the prompt ``I like Italy and France. I visited the city of''. This model responds with ``Rome''. We keep the target token ``Paris'' and compute the loss to obtain the RA maps, presented in Figure \ref{fig: opt350m norms i like}.

This model has 384 attention heads. We sort the heads according to their RA maps' norms and present the one with the highest norm in \autoref{fig: opt350m forward and reversed attn i like}, revealing a pattern similar to that of GPT2-xl in \autoref{fig: gpt2xl forward and reversed attn i like}.

To emphasize that heads with low RA norm are barely updated, we maintain the same coloring scale from the head with the highest norm and present the 11-th highest by norm head. This head's RA map appears empty, indicating all scores are close to zero. It is also evident that the forward pass attention map of this head only attends to the first token of the sentence, a phenomenon known as ``attention sink'' \citep{xiao2023efficient}. This seems to be a default pattern when the function that the head describes is not activated.

\clearpage

\begin{figure*}
    \centering
    \begin{subfigure}[h]{0.98\textwidth}
        \centering
        \includegraphics[width=\linewidth]{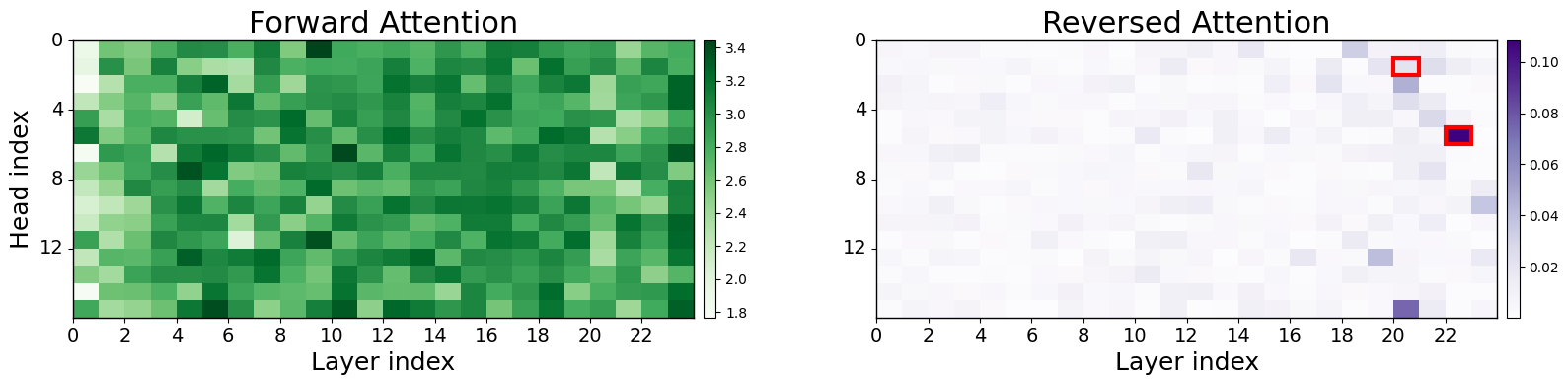}
        \caption{Attention heads by norm}
    \end{subfigure}

    \begin{subfigure}[h]{0.95\textwidth}
        \centering
        \includegraphics[width=\linewidth]{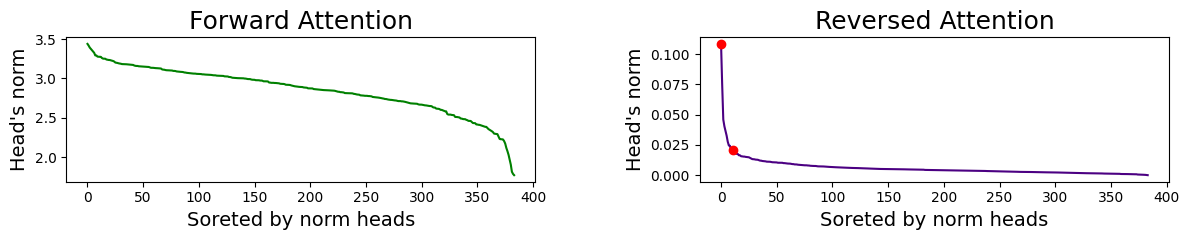}
        \caption{Sorted by attention heads norm values}
    \end{subfigure}
    
    \caption{The forward and Reversed Attention (RA) of OPT-350m, given the editing target ``Paris'' and the prompt ``I like Italy and France. I visited the city of''.
      The head with the highest RA norm is highlighted in red as well as the 11-th highest.
    }
    \label{fig: opt350m norms i like}
\end{figure*}

\begin{figure*}
    \centering
    \begin{subfigure}[h]{0.37\textwidth}
        \centering
        \includegraphics[width=\linewidth]{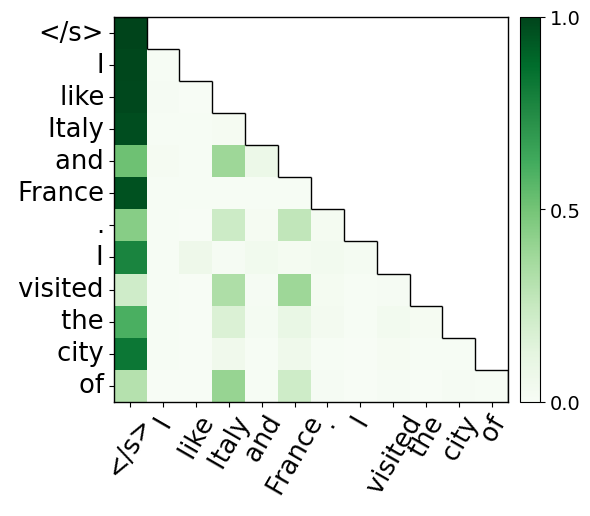}
        \caption{L22H5 - Forward}
    \end{subfigure}
    \hspace{1.5cm}
    \begin{subfigure}[h]{0.37\textwidth}
        \centering
        \includegraphics[width=\linewidth]{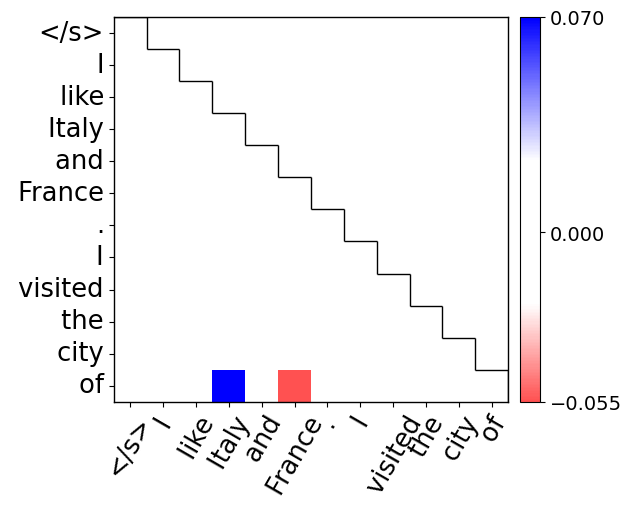}
        \caption{L22H5 - Reversed}
        \label{subfig: opt Reversed - Paris - highest}
    \end{subfigure}
    
    \begin{subfigure}[h]{0.37\textwidth}
        \centering
        \includegraphics[width=\linewidth]{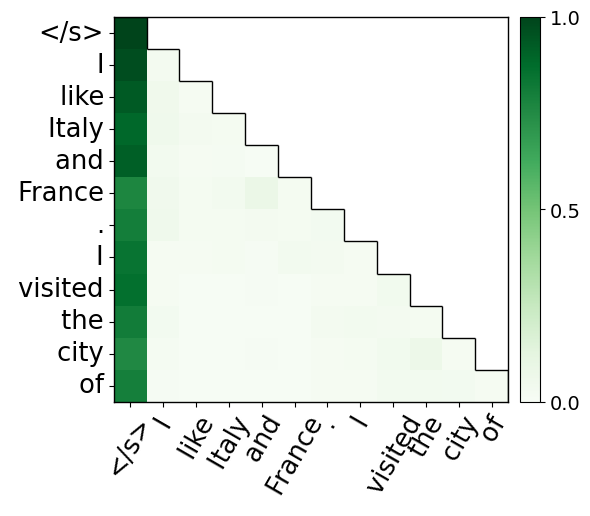}
        \caption{L20H1 - Forward}
    \end{subfigure}
    \hspace{1.5cm}
    \begin{subfigure}[h]{0.37\textwidth}
        \centering
        \includegraphics[width=\linewidth]{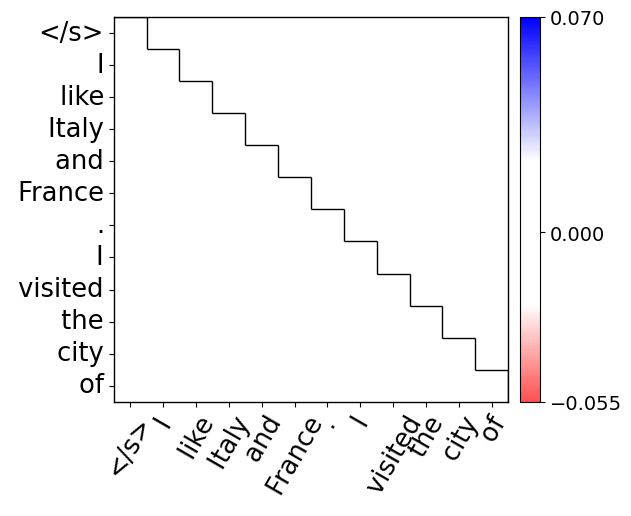}
        \caption{L20H1 - Reversed}
        \label{subfig: opt Reversed - Paris - mid}
    \end{subfigure}
    \caption{Attention maps for the prompt ``I like Italy and France. I visited the city of''.
    The head with the highest Reversed Attention (RA) norm is head 5 at layer 22 (L22H5). Its RA map (b) shows how it amplifies the attention score for ``France'' while reducing that of ``Italy''.
    Head 1 at layer 20 is the 11-th highest head by RA norm. We present its RA map (d) using the same color bar scale as that of L22H5, which highlights that this head barely undergoes any update.
    }
    \label{fig: opt350m forward and reversed attn i like}
\end{figure*}

\clearpage
\clearpage

\section{Perturbation Test}
\label{Perturbation test}
In this section, we present the full results of the perturbation analysis outlined in \autoref{Reversed Attention identifies attention heads' importance}.
This analysis can be viewed as a sparsity test, wherein we mask (zero-out) all attention heads and gradually unmask them according to different methods.

Throughout this section, we examine the ordering provided by Reversed Attention (RA) norms and by Causal Mediation (CM).
As a baseline, we also consider the ordering provided by the norm of the average forward pass attention scores (``Forward''), random order (``Random''), and the order of the attention heads in the model (``Index'').
Additionally, we evaluate each method by the order it provides and by the reverse order, denoted by the suffix ``$[\leftarrow$]''.
For example, RA orders the heads from ones with the largest norm to the one with the smallest, while ``RA$[\leftarrow$]'' orders them from the smallest to the largest.
This reversal in orders represents a negative perturbation test, whereas the original order represents a positive test.
A successful order in such a case should yield high Area Under the Curve (AUC) in the positive test and low AUC in the negative test, for the graph that measures the model's accuracy as we unmask the heads in the order of the method.

\paragraph{Casual Mediation implementation}
CM is applied separately for each task, returning the average indirect effect (AIE) for each attention head.
For all tasks, we exclude the AIE when applying CM on the prompts' last tokens (the target token), as it is the token that the model should edit.
One implementation we use (CM1) zeros out each head separately.

Another implementation we examined (CM2) is the implementation provided by \citep{todd2023function}. This implementation instead of zeroing-out the heads, uses the average activation of each head, as collected from a few left-out examples. 
An example for the CM2 we produced, compare to the RA norms, is shown in \autoref{fig: RA vs CM}.

\begin{figure}[h!]
    \centering
    \begin{subfigure}[h]{0.47\textwidth}
        \centering
        \includegraphics[width=\linewidth]{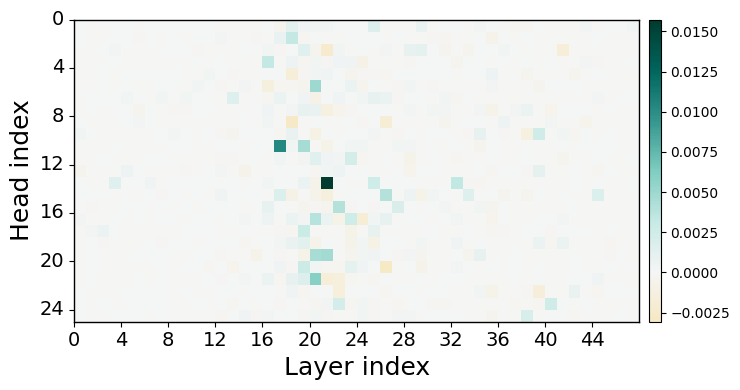} 
        \caption{Casual mediation (AIE) implementation by \citet{todd2023function}}
        \label{subfig: cm map}
    \end{subfigure}
    \hfill
    \begin{subfigure}[h]{0.47\textwidth}
        \centering
        \includegraphics[width=\linewidth]{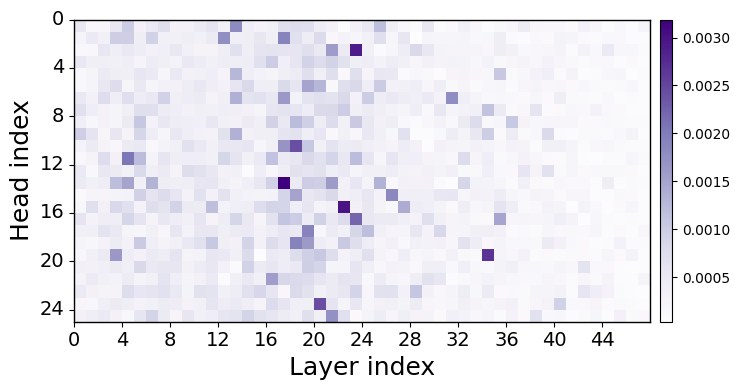} 
        \caption{Reversed Attention (norm)}
        \label{subfig: RA map}
    \end{subfigure}
    \caption{GPT2-xl casual mediation and Reversed Attention maps for the ICL capitalize task with 5-shots.}
    \label{fig: RA vs CM}
\end{figure}

\paragraph{Datasets}
\label{Datasets}
The ICL tasks we used are sourced from \citep{todd2023function}.
Each task consists of a set of pairs with a common relation in the format ``<question>,<answer>'', where the model is tasked with completing the answer given the question.
For constructing the prompt for each task and pair, we used the following format: ``Q: <question> \textbackslash A: <answer> \textbackslash\textbackslash''.
For example, a 2-shot prompt is in the form of: ``Q: <question1> \textbackslash A: <answer1> \textbackslash\textbackslash Q: <question2> \textbackslash A: <answer2> \textbackslash\textbackslash Q: <question> \textbackslash A:''.

The natural language questions we utilized are from \citep{hernandez2024linearity}.
This dataset comprises tasks, each with a set of pairs in the format ``<question>,<answer>'', along with a natural language format for constructing the prompt.
For instance, in the task concerning the relationship between persons that plays pro sports, the format is ``<question> plays the sport of <answer>''. 
We observed that two tasks, country-capital and present-past, had too few examples but shared the same relation as tasks in the ICL dataset by \citep{todd2023function}.\footnote{The Antonym and Synonym tasks used by \citep{todd2023function} are from \citep{nguyen2017distinguishing} and the English-French task is from \citep{lample2018word}.}
We employed the format from \citep{hernandez2024linearity} along with the pairs from \citep{todd2023function}.

In practice, we additionally examined the natural language tasks by prompting the tasks' question with few-shots labeled example.
Hence, the 0-shots represent the original tasks, while the n-shots results are provided for comprehensive examination. We adopt this practice from NLP benchmark like HuggingFace leaderborad.\footnote{ \url{https://huggingface.co/spaces/open-llm-leaderboard-old/open\_llm\_leaderboard}}
For example, a common practice is to evaluate HellaSwag \citep{zellers2019hellaswag} with 10-shots and MMLU \citep{hendrycksmeasuring} with 5-shots.

The full list of tasks is provided in \autoref{tab: datasets list}.
To summarize, we have ICL tasks with a uniform format of prompt, where the model is required to infer the relation between the question and the answer from given examples.
Additionally, we have natural language questions (with and without given examples) where the relation between the question and the answer is explicitly stated in the prompt.
In all experiments, and for each task separately, we used a split of $1/3$ from all available examples to extract the test set. This set of examples is used to report our results and is not included in the creation of any method (i.e. RA, CM).

\begin{table*}
  \caption{List of tasks}
  \label{tab: datasets list}
  \centering
  \begin{small}
  \begin{tabular}{ccccc}
    \toprule
    Task     &  Task source and type   & Examples \\
    \midrule 
adjective-v-verb-3 & \multirow{18}{11em}{ ICL \citep{todd2023function}} & Q: uplifting, approve, decide A: [uplifting] &  \\ 
alphabetically-first-3 &     & Q: blissful, rat, dingo A: [blissful] &  \\ 
animal-v-object-3 &     & Q: bicycle, skunk, egg A: [skunk] &  \\ 
antonym &     & Q: expire A: [renew] &  \\ 
capitalize &     & Q: cow A: [Cow] &  \\ 
capitalize-first-letter &     & Q: bunny A: [B] &  \\ 
choose-middle-of-3 &     & Q: white, house, wallet A: [house] &  \\ 
country-capital &     & Q: Cabo Verde A: [Praia] &  \\ 
english-french &     & Q: careful A: [prudent] &  \\ 
next-item &     & Q: 12 A: [13] &  \\ 
person-sport &     & Q: Lou Gehrig A: [baseball] &  \\ 
present-past &     & Q: justify A: [justified] &  \\ 
prev-item &     & Q: 13 A: [12] &  \\ 
singular-plural &     & Q: glue A: [glues] &  \\ 
synonym &     & Q: missing A: [lost] &  \\ 
    \midrule 
company-hq & \multirow{5}{11em}{ Natural question \citep{hernandez2024linearity}} & EMI is headquartered in the city of [London] &  \\ 
landmark-in-country &          & route 75 is in the country of [Australia] &  \\ 
person-plays-pro-sport &          & Lou Gehrig plays the sport of [baseball] &  \\ 
product-by-company &          & Digital Negative was created by [Adobe] &  \\ 
  \midrule 
country-capital & \multirow{2}{11em}{ Natural questoin \citep{hernandez2024linearity}, \citep{todd2023function}} & The capital city of Cabo Verde is [Praia] &  \\ 
present-past &          & The past tense of justify is [justified] &  \\ 
\\
    
    \bottomrule
  \end{tabular}
  \end{small}
\end{table*}

\begin{figure*}
    \centering
    \begin{subfigure}[h]{0.57\textwidth}
        \centering
        \includegraphics[width=\linewidth]{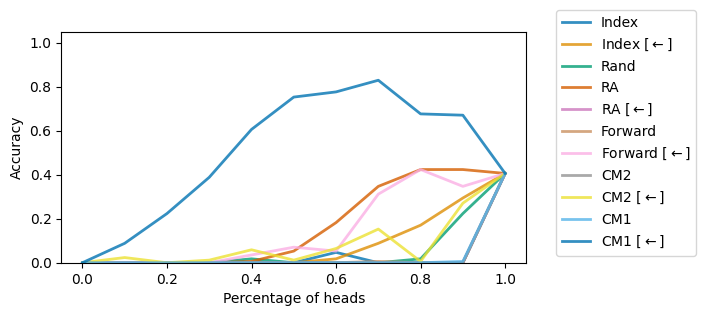}
        \caption{0-shots}
        \label{subfig: AUC bar}
    \end{subfigure}
    \hfill
    \begin{subfigure}[h]{0.42\textwidth}
        \centering
        \includegraphics[width=\linewidth]{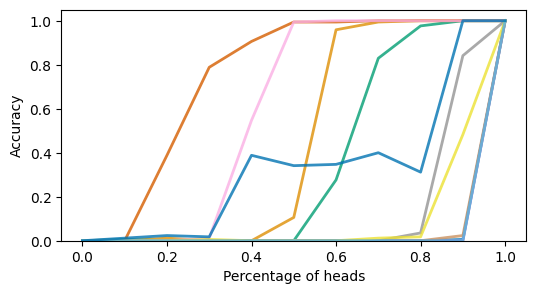}
        \caption{10-shots}
        \label{subfig: AUC graph}
    \end{subfigure}
    
    \caption{Perturbation test results visualized for Llama2-7B on the capitalize ICL task. With no shots, when the original model achieves only 0.4 accuracy, causal mediation (CM) achieve the highest AUC score. With few-shots prompting, Reversed Attention (RA) outperforms all other methods.
    }
    \label{fig: Perturbation viz}
\end{figure*}

\subsection{Perturbation results}
We conducted the perturbation test on GPT2-xl \cite{Radford2019LanguageMA}, OPT-1.3B \cite{zhang2022opt}, GPT-j \cite{gpt-j} and Llama2-7B \cite{touvron2023llama} models.
For each task and method, we provided 25 examples to extract the order of the attention heads.
We quantified the performance of each method by the AUC of the graph that measures the model's accuracy as we unmask the heads.
\autoref{fig: Perturbation viz} provides an example of such a graph and the AUC results extracted from it.

The full ICL results are displayed in \autoref{tab: perturbation appendix gpt2xl icl} ~\ref{tab: perturbation appendix opt1.3B icl} ~\ref{tab: perturbation appendix gptj icl} and ~\ref{tab: perturbation appendix llama2-7B icl}.
We notice that for all models, RA achieves the best results in the majority of the tasks when prompted with 5 or 10 shots, only falling behind CM without any shots.
It is also evident that RA shows more dominance across tasks with relatively larger models, like LLaMA2-7B.

The natural language task results are provided in \autoref{tab: perturbation appendix gpt2xl natural} ~\ref{tab: perturbation appendix opt1.3 natural} ~\ref{tab: perturbation appendix gptj natural} and ~\ref{tab: perturbation appendix llama2-7B natural}. 
When considering the 0-shot setting, which is equivalent to examining the perturbation task on the original natural language tasks, we find that when the models' original ability to answer the asked question is high, RA achieves better results compare to CM.
In all other cases CM is better only under the 0-shot setup. With few-shot prompting, RA is the preferred method.

In summary, the perturbation test reveals that RA has the potential to localize the model's behavior as good as existing methods and in particular in ICL.

\begin{table*}
  \centering
  \begin{small}
  \begin{tabular}{@{}c@{~~}c@{~~}c@{~~}c@{~~}c@{~~}c@{~~}c@{~~}c@{~~}c@{~~}c@{~~}c@{~~}c@{~~}c@{~~}}
    \toprule
     
    Task   & n-shots & Random & Index [$\leftarrow$] & Index & CM2 [$\leftarrow$] & CM2 & CM1 [$\leftarrow$] & CM1 & FA [$\leftarrow$] & FA & RA [$\leftarrow$] & RA \\
    \midrule

\multirow{4}{3em}{adjective-v-verb-3}  & 0 & 0.03 & 0.04 & 0.05 & 0.09 & 0.01 & \textbf{0.44} & 0.01 & 0.03 & 0.03 & 0.01 & 0.05 \\ 
 & 1 & 0.21 & 0.18 & 0.09 & 0.27 & 0.03 & \textbf{0.63} & 0.02 & 0.1 & 0.12 & 0.03 & 0.3 \\ 
 & 5 & 0.23 & 0.22 & 0.08 & 0.21 & 0.05 & 0.15 & 0.03 & 0.18 & 0.14 & 0.03 & \textbf{0.35} \\ 
 & 10 & 0.25 & 0.24 & 0.09 & 0.12 & 0.1 & \textbf{0.4} & 0.04 & 0.2 & 0.11 & 0.03 & 0.36 \\ 
\multirow{4}{3em}{alphabetic-ally-first-3}  & 0 & 0.02 & 0.03 & 0.05 & 0.15 & 0.01 & \textbf{0.28} & 0.01 & 0.02 & 0.04 & 0.02 & 0.04 \\ 
 & 1 & 0.12 & 0.13 & 0.06 & 0.07 & 0.11 & \textbf{0.22} & 0.02 & 0.07 & 0.08 & 0.02 & 0.2 \\ 
 & 5 & 0.15 & 0.15 & 0.05 & 0.08 & 0.11 & 0.19 & 0.05 & 0.1 & 0.08 & 0.02 & \textbf{0.23} \\ 
 & 10 & 0.14 & 0.13 & 0.05 & 0.11 & 0.13 & \textbf{0.23} & 0.05 & 0.1 & 0.06 & 0.02 & 0.22 \\ 
\multirow{4}{3em}{animal-v-object-3}  & 0 & 0.09 & 0.03 & 0.06 & 0.11 & 0.01 & \textbf{0.28} & 0.01 & 0.03 & 0.04 & 0.02 & 0.08 \\ 
 & 1 & 0.2 & 0.16 & 0.08 & 0.3 & 0.04 & \textbf{0.39} & 0.03 & 0.13 & 0.11 & 0.02 & 0.29 \\ 
 & 5 & 0.28 & 0.27 & 0.1 & 0.18 & 0.13 & \textbf{0.44} & 0.04 & 0.24 & 0.16 & 0.04 & 0.42 \\ 
 & 10 & 0.3 & 0.29 & 0.1 & 0.23 & 0.1 & 0.3 & 0.05 & 0.27 & 0.15 & 0.05 & \textbf{0.49} \\ 
\multirow{4}{3em}{antonym}  & 0 & 0.0 & 0.0 & 0.0 & 0.0 & 0.0 & \textbf{0.01} & 0.0 & 0.0 & 0.0 & 0.0 & 0.0 \\ 
 & 1 & 0.02 & 0.07 & 0.02 & 0.01 & 0.01 & \textbf{0.15} & 0.01 & 0.02 & 0.02 & 0.01 & 0.07 \\ 
 & 5 & 0.09 & 0.24 & 0.06 & 0.03 & 0.03 & 0.27 & 0.03 & 0.15 & 0.05 & 0.02 & \textbf{0.32} \\ 
 & 10 & 0.11 & 0.26 & 0.07 & 0.03 & 0.03 & 0.3 & 0.03 & 0.17 & 0.05 & 0.03 & \textbf{0.34} \\ 
\multirow{4}{3em}{capitalize}  & 0 & 0.01 & 0.0 & 0.0 & 0.08 & 0.0 & \textbf{0.23} & 0.0 & 0.0 & 0.0 & 0.03 & 0.0 \\ 
 & 1 & 0.08 & 0.19 & 0.05 & 0.12 & 0.05 & \textbf{0.64} & 0.01 & 0.03 & 0.07 & 0.03 & 0.1 \\ 
 & 5 & 0.31 & 0.43 & 0.14 & 0.1 & 0.27 & \textbf{0.52} & 0.05 & 0.26 & 0.17 & 0.05 & 0.5 \\ 
 & 10 & 0.36 & 0.45 & 0.14 & 0.08 & 0.27 & \textbf{0.59} & 0.05 & 0.32 & 0.16 & 0.05 & \textbf{0.59} \\ 
\multirow{4}{3em}{choose-middle-of-3}  & 0 & 0.02 & 0.02 & 0.02 & 0.05 & 0.01 & \textbf{0.24} & 0.01 & 0.01 & 0.02 & 0.01 & 0.02 \\ 
 & 1 & 0.1 & 0.16 & 0.06 & 0.1 & 0.05 & \textbf{0.35} & 0.03 & 0.05 & 0.08 & 0.02 & 0.19 \\ 
 & 5 & 0.17 & 0.23 & 0.07 & 0.04 & 0.11 & 0.21 & 0.05 & 0.13 & 0.1 & 0.03 & \textbf{0.3} \\ 
 & 10 & 0.2 & 0.25 & 0.07 & 0.05 & 0.25 & 0.2 & 0.05 & 0.14 & 0.08 & 0.04 & \textbf{0.35} \\ 
\multirow{4}{3em}{country-capital}  & 0 & 0.0 & 0.0 & 0.01 & 0.02 & 0.0 & \textbf{0.04} & 0.0 & 0.0 & 0.02 & 0.0 & 0.0 \\ 
 & 1 & 0.07 & 0.18 & 0.08 & 0.08 & 0.02 & 0.22 & 0.02 & 0.08 & 0.09 & 0.03 & \textbf{0.29} \\ 
 & 5 & 0.19 & 0.3 & 0.1 & 0.05 & 0.05 & 0.4 & 0.04 & 0.24 & 0.07 & 0.04 & \textbf{0.42} \\ 
 & 10 & 0.19 & 0.29 & 0.1 & 0.07 & 0.08 & \textbf{0.55} & 0.03 & 0.24 & 0.07 & 0.04 & 0.4 \\ 
\multirow{4}{3em}{english-french}  & 0 & 0.0 & 0.0 & 0.0 & 0.0 & 0.0 & \textbf{0.01} & 0.0 & 0.0 & 0.0 & 0.0 & 0.0 \\ 
 & 1 & 0.01 & 0.03 & 0.01 & 0.01 & 0.01 & 0.02 & 0.0 & 0.02 & 0.01 & 0.01 & \textbf{0.04} \\ 
 & 5 & 0.06 & 0.08 & 0.02 & 0.02 & 0.03 & 0.08 & 0.01 & 0.06 & 0.03 & 0.01 & \textbf{0.12} \\ 
 & 10 & 0.07 & 0.09 & 0.03 & 0.02 & 0.04 & 0.06 & 0.02 & 0.07 & 0.03 & 0.01 & \textbf{0.13} \\ 
\multirow{4}{3em}{lowercase-last-letter}  & 0 & 0.0 & 0.0 & 0.0 & 0.0 & 0.0 & 0.0 & 0.0 & 0.0 & 0.0 & 0.0 & 0.0 \\ 
 & 1 & 0.02 & 0.03 & 0.01 & 0.03 & 0.02 & \textbf{0.08} & 0.0 & 0.01 & 0.02 & 0.01 & 0.06 \\ 
 & 5 & 0.05 & 0.06 & 0.02 & 0.06 & 0.04 & \textbf{0.14} & 0.01 & 0.04 & 0.02 & 0.01 & 0.1 \\ 
 & 10 & 0.06 & 0.08 & 0.02 & 0.07 & 0.04 & \textbf{0.11} & 0.03 & 0.05 & 0.03 & 0.01 & 0.1 \\ 
\multirow{4}{3em}{next-item}  & 0 & 0.02 & 0.01 & 0.03 & 0.06 & 0.0 & \textbf{0.22} & 0.0 & 0.02 & 0.01 & 0.01 & 0.03 \\ 
 & 1 & 0.06 & 0.1 & 0.04 & 0.07 & 0.02 & \textbf{0.25} & 0.01 & 0.05 & 0.04 & 0.03 & 0.14 \\ 
 & 5 & 0.17 & 0.26 & 0.08 & 0.1 & 0.05 & 0.31 & 0.03 & 0.2 & 0.07 & 0.03 & \textbf{0.34} \\ 
 & 10 & 0.22 & 0.27 & 0.09 & 0.12 & 0.09 & 0.3 & 0.04 & 0.21 & 0.08 & 0.04 & \textbf{0.4} \\ 
\multirow{4}{3em}{person-sport}  & 0 & 0.0 & 0.0 & 0.0 & 0.0 & 0.0 & 0.0 & 0.0 & 0.0 & 0.0 & 0.0 & 0.0 \\ 
 & 1 & 0.22 & 0.33 & 0.09 & 0.18 & 0.13 & 0.35 & 0.03 & 0.11 & 0.1 & 0.06 & \textbf{0.37} \\ 
 & 5 & 0.21 & 0.34 & 0.09 & 0.18 & 0.16 & 0.34 & 0.07 & 0.24 & 0.09 & 0.09 & \textbf{0.44} \\ 
 & 10 & 0.22 & 0.33 & 0.09 & 0.23 & 0.19 & 0.31 & 0.11 & 0.31 & 0.11 & 0.09 & \textbf{0.45} \\ 
\multirow{4}{3em}{present-past}  & 0 & 0.0 & 0.0 & 0.0 & 0.0 & 0.0 & \textbf{0.02} & 0.0 & 0.0 & 0.0 & 0.0 & 0.0 \\ 
 & 1 & 0.05 & 0.19 & 0.05 & 0.1 & 0.05 & \textbf{0.45} & 0.02 & 0.07 & 0.05 & 0.02 & 0.13 \\ 
 & 5 & 0.34 & 0.49 & 0.15 & 0.07 & 0.27 & \textbf{0.68} & 0.06 & 0.34 & 0.19 & 0.05 & 0.65 \\ 
 & 10 & 0.36 & 0.49 & 0.15 & 0.06 & 0.27 & \textbf{0.77} & 0.05 & 0.41 & 0.19 & 0.05 & 0.69 \\ 
\multirow{4}{3em}{prev-item}  & 0 & 0.01 & 0.01 & 0.03 & 0.06 & 0.0 & \textbf{0.12} & 0.0 & 0.01 & 0.02 & 0.01 & 0.03 \\ 
 & 1 & 0.04 & 0.07 & 0.02 & 0.06 & 0.02 & \textbf{0.26} & 0.01 & 0.03 & 0.02 & 0.02 & 0.09 \\ 
 & 5 & 0.08 & 0.17 & 0.05 & 0.08 & 0.04 & 0.24 & 0.02 & 0.12 & 0.05 & 0.02 & \textbf{0.25} \\ 
 & 10 & 0.1 & 0.17 & 0.07 & 0.07 & 0.03 & 0.23 & 0.02 & 0.14 & 0.05 & 0.03 & \textbf{0.25} \\ 
\multirow{4}{3em}{singular-plural}  & 0 & 0.01 & 0.02 & 0.02 & 0.02 & 0.01 & \textbf{0.2} & 0.0 & 0.0 & 0.02 & 0.03 & 0.01 \\ 
 & 1 & 0.17 & 0.21 & 0.07 & 0.13 & 0.07 & \textbf{0.48} & 0.02 & 0.1 & 0.09 & 0.06 & 0.24 \\ 
 & 5 & 0.38 & 0.43 & 0.14 & 0.14 & 0.38 & 0.51 & 0.09 & 0.35 & 0.2 & 0.06 & \textbf{0.64} \\ 
 & 10 & 0.4 & 0.45 & 0.13 & 0.13 & 0.36 & 0.44 & 0.11 & 0.41 & 0.18 & 0.06 & \textbf{0.58} \\ 
\multirow{4}{3em}{synonym}  & 0 & 0.0 & 0.0 & 0.0 & 0.0 & 0.0 & \textbf{0.01} & 0.0 & 0.0 & 0.0 & 0.0 & 0.0 \\ 
 & 1 & 0.01 & 0.01 & 0.0 & 0.0 & 0.0 & \textbf{0.04} & 0.0 & 0.01 & 0.0 & 0.0 & 0.02 \\ 
 & 5 & 0.01 & 0.01 & 0.0 & 0.01 & 0.0 & \textbf{0.04} & 0.0 & 0.01 & 0.0 & 0.0 & 0.03 \\ 
 & 10 & 0.01 & 0.01 & 0.0 & 0.0 & 0.0 & \textbf{0.03} & 0.0 & 0.01 & 0.0 & 0.0 & \textbf{0.03} \\ 

    \bottomrule
  \end{tabular}
  \end{small}
  \caption{GPT2-xl perturbation test on ICL tasks. Each score represents the AUC with respect to n-shot and a method that orders all the attention heads. The methods include random ordering, Index (from the first layer to the last), Causal Mediation (CM, two variations of implementation), forward attention norm (FA), and Reversed Attention norm (RA). For each ordering method, we also examine the reversed order, annotated by [$\leftarrow$].}
  \label{tab: perturbation appendix gpt2xl icl}
\end{table*}

\begin{table*}
  \centering
  \begin{small}
  \begin{tabular}{@{}c@{~~}c@{~~}c@{~~}c@{~~}c@{~~}c@{~~}c@{~~}c@{~~}c@{~~}c@{~~}c@{~~}c@{~~}c@{~~}}
    \toprule

    Task   & n-shots & Random & Index [$\leftarrow$] & Index & CM2 [$\leftarrow$] & CM2 & CM1 [$\leftarrow$] & CM1 & FA [$\leftarrow$] & FA & RA [$\leftarrow$] & RA \\
    \midrule

\multirow{4}{3em}{adjective-v-verb-3}  & 0 & 0.01 & 0.01 & 0.03 & 0.03 & 0.0 & \textbf{0.29} & 0.0 & 0.01 & 0.06 & 0.01 & 0.02 \\ 
 & 1 & 0.18 & 0.1 & 0.04 & 0.24 & 0.02 & \textbf{0.42} & 0.02 & 0.12 & 0.15 & 0.02 & 0.39 \\ 
 & 5 & 0.24 & 0.11 & 0.05 & 0.08 & 0.08 & 0.11 & 0.03 & 0.2 & 0.13 & 0.02 & \textbf{0.45} \\ 
 & 10 & 0.24 & 0.11 & 0.03 & 0.02 & 0.11 & 0.03 & 0.06 & 0.17 & 0.11 & 0.02 & \textbf{0.34} \\ 
\multirow{4}{3em}{alphabetic-ally-first-3}  & 0 & 0.01 & 0.01 & 0.03 & 0.1 & 0.01 & \textbf{0.23} & 0.01 & 0.02 & 0.06 & 0.01 & 0.05 \\ 
 & 1 & 0.09 & 0.08 & 0.03 & 0.04 & 0.07 & 0.16 & 0.02 & 0.09 & 0.1 & 0.02 & \textbf{0.24} \\ 
 & 5 & 0.12 & 0.08 & 0.03 & 0.05 & 0.12 & 0.06 & 0.07 & 0.13 & 0.08 & 0.02 & \textbf{0.25} \\ 
 & 10 & 0.12 & 0.08 & 0.02 & 0.03 & 0.09 & \textbf{0.22} & 0.01 & 0.11 & 0.06 & 0.02 & 0.2 \\ 
\multirow{4}{3em}{animal-v-object-3}  & 0 & 0.02 & 0.01 & 0.03 & 0.01 & 0.01 & \textbf{0.31} & 0.01 & 0.03 & 0.06 & 0.01 & 0.06 \\ 
 & 1 & 0.16 & 0.06 & 0.04 & 0.1 & 0.06 & \textbf{0.55} & 0.02 & 0.11 & 0.11 & 0.02 & 0.29 \\ 
 & 5 & 0.21 & 0.1 & 0.04 & 0.06 & 0.07 & \textbf{0.39} & 0.03 & 0.18 & 0.14 & 0.03 & 0.36 \\ 
 & 10 & 0.19 & 0.1 & 0.02 & 0.06 & 0.1 & 0.09 & 0.07 & 0.18 & 0.12 & 0.03 & \textbf{0.31} \\ 
\multirow{4}{3em}{antonym}  & 0 & 0.0 & 0.0 & 0.0 & 0.0 & 0.0 & \textbf{0.16} & 0.0 & 0.0 & 0.0 & 0.0 & 0.01 \\ 
 & 1 & 0.03 & 0.04 & 0.01 & 0.02 & 0.01 & \textbf{0.19} & 0.01 & 0.05 & 0.02 & 0.01 & 0.09 \\ 
 & 5 & 0.07 & 0.09 & 0.02 & 0.03 & 0.02 & 0.1 & 0.02 & 0.12 & 0.07 & 0.02 & \textbf{0.21} \\ 
 & 10 & 0.08 & 0.1 & 0.02 & 0.02 & 0.02 & 0.05 & 0.02 & 0.12 & 0.06 & 0.02 & \textbf{0.2} \\ 
\multirow{4}{3em}{capitalize}  & 0 & 0.02 & 0.0 & 0.0 & 0.04 & 0.0 & \textbf{0.29} & 0.0 & 0.02 & 0.01 & 0.01 & 0.02 \\ 
 & 1 & 0.03 & 0.02 & 0.01 & 0.01 & 0.0 & \textbf{0.19} & 0.0 & 0.01 & 0.01 & 0.0 & 0.05 \\ 
 & 5 & 0.35 & 0.24 & 0.09 & 0.08 & 0.23 & 0.52 & 0.05 & 0.38 & 0.19 & 0.05 & \textbf{0.67} \\ 
 & 10 & 0.36 & 0.24 & 0.06 & 0.07 & 0.23 & 0.47 & 0.05 & 0.39 & 0.18 & 0.05 & \textbf{0.6} \\ 
\multirow{4}{3em}{choose-middle-of-3}  & 0 & 0.01 & 0.0 & 0.0 & 0.02 & 0.0 & \textbf{0.14} & 0.0 & 0.01 & 0.02 & 0.0 & 0.01 \\ 
 & 1 & 0.05 & 0.05 & 0.02 & 0.02 & 0.04 & 0.14 & 0.01 & 0.07 & 0.07 & 0.01 & \textbf{0.18} \\ 
 & 5 & 0.13 & 0.12 & 0.04 & 0.04 & 0.13 & 0.08 & 0.04 & 0.19 & 0.11 & 0.03 & \textbf{0.31} \\ 
 & 10 & 0.14 & 0.11 & 0.03 & 0.03 & 0.1 & 0.11 & 0.04 & 0.16 & 0.07 & 0.03 & \textbf{0.25} \\ 
\multirow{4}{3em}{country-capital}  & 0 & 0.0 & 0.0 & 0.0 & 0.0 & 0.0 & \textbf{0.04} & 0.0 & 0.01 & 0.0 & 0.0 & 0.02 \\ 
 & 1 & 0.15 & 0.14 & 0.05 & 0.04 & 0.12 & \textbf{0.52} & 0.04 & 0.22 & 0.11 & 0.03 & 0.43 \\ 
 & 5 & 0.22 & 0.19 & 0.07 & 0.06 & 0.05 & 0.55 & 0.05 & 0.28 & 0.19 & 0.04 & \textbf{0.56} \\ 
 & 10 & 0.2 & 0.17 & 0.04 & 0.06 & 0.08 & 0.38 & 0.05 & 0.28 & 0.13 & 0.04 & \textbf{0.46} \\ 
\multirow{4}{3em}{english-french}  & 0 & 0.0 & 0.0 & 0.0 & 0.0 & 0.0 & \textbf{0.02} & 0.0 & 0.0 & 0.0 & 0.0 & 0.0 \\ 
 & 1 & 0.02 & 0.03 & 0.01 & 0.01 & 0.01 & 0.03 & 0.01 & 0.03 & 0.01 & 0.0 & \textbf{0.07} \\ 
 & 5 & 0.1 & 0.08 & 0.02 & 0.02 & 0.04 & 0.16 & 0.02 & 0.1 & 0.05 & 0.02 & \textbf{0.19} \\ 
 & 10 & 0.11 & 0.08 & 0.02 & 0.02 & 0.05 & 0.08 & 0.02 & 0.11 & 0.05 & 0.02 & \textbf{0.18} \\ 
\multirow{4}{3em}{lowercase-last-letter}  & 0 & 0.0 & 0.0 & 0.0 & 0.0 & 0.0 & 0.0 & 0.0 & 0.0 & 0.0 & 0.0 & 0.0 \\ 
 & 1 & 0.02 & 0.02 & 0.01 & 0.01 & 0.01 & \textbf{0.07} & 0.01 & 0.01 & 0.01 & 0.0 & \textbf{0.07} \\ 
 & 5 & 0.04 & 0.03 & 0.01 & 0.06 & 0.04 & 0.02 & 0.02 & 0.04 & 0.02 & 0.01 & \textbf{0.08} \\ 
 & 10 & 0.04 & 0.03 & 0.01 & \textbf{0.08} & 0.03 & 0.02 & 0.04 & 0.05 & 0.03 & 0.01 & \textbf{0.08} \\ 
\multirow{4}{3em}{next-item}  & 0 & 0.01 & 0.02 & 0.01 & 0.1 & 0.01 & \textbf{0.27} & 0.01 & 0.04 & 0.04 & 0.01 & 0.07 \\ 
 & 1 & 0.08 & 0.08 & 0.02 & 0.09 & 0.02 & \textbf{0.35} & 0.02 & 0.09 & 0.08 & 0.02 & 0.25 \\ 
 & 5 & 0.15 & 0.12 & 0.03 & 0.05 & 0.03 & 0.34 & 0.03 & 0.18 & 0.11 & 0.03 & \textbf{0.41} \\ 
 & 10 & 0.14 & 0.14 & 0.03 & 0.06 & 0.04 & 0.2 & 0.03 & 0.2 & 0.1 & 0.03 & \textbf{0.39} \\ 
\multirow{4}{3em}{person-sport}  & 0 & 0.0 & 0.0 & 0.0 & 0.0 & 0.0 & 0.0 & 0.0 & 0.0 & 0.0 & 0.0 & 0.0 \\ 
 & 1 & 0.18 & 0.17 & 0.05 & 0.07 & 0.12 & \textbf{0.46} & 0.03 & 0.17 & 0.11 & 0.04 & \textbf{0.46} \\ 
 & 5 & 0.27 & 0.2 & 0.06 & 0.09 & 0.2 & \textbf{0.5} & 0.06 & 0.34 & 0.16 & 0.04 & 0.49 \\ 
 & 10 & 0.23 & 0.22 & 0.04 & 0.21 & 0.18 & 0.25 & 0.09 & 0.36 & 0.12 & 0.06 & \textbf{0.38} \\ 
\multirow{4}{3em}{present-past}  & 0 & 0.0 & 0.0 & 0.0 & \textbf{0.01} & 0.0 & \textbf{0.01} & 0.0 & \textbf{0.01} & \textbf{0.01} & 0.0 & \textbf{0.01} \\ 
 & 1 & 0.07 & 0.04 & 0.01 & 0.03 & 0.02 & 0.07 & 0.0 & 0.05 & 0.04 & 0.01 & \textbf{0.09} \\ 
 & 5 & 0.25 & 0.23 & 0.06 & 0.05 & 0.25 & 0.27 & 0.04 & 0.37 & 0.19 & 0.04 & \textbf{0.48} \\ 
 & 10 & 0.3 & 0.27 & 0.07 & 0.13 & 0.24 & 0.44 & 0.05 & 0.41 & 0.2 & 0.05 & \textbf{0.61} \\ 
\multirow{4}{3em}{prev-item}  & 0 & 0.01 & 0.01 & 0.01 & 0.02 & 0.0 & \textbf{0.16} & 0.0 & 0.02 & 0.03 & 0.0 & 0.05 \\ 
 & 1 & 0.04 & 0.04 & 0.01 & 0.07 & 0.01 & \textbf{0.17} & 0.01 & 0.04 & 0.03 & 0.01 & 0.1 \\ 
 & 5 & 0.09 & 0.07 & 0.02 & 0.07 & 0.03 & 0.17 & 0.01 & 0.1 & 0.07 & 0.02 & \textbf{0.2} \\ 
 & 10 & 0.11 & 0.1 & 0.02 & 0.02 & 0.03 & 0.09 & 0.02 & 0.13 & 0.07 & 0.02 & \textbf{0.24} \\ 
\multirow{4}{3em}{singular-plural}  & 0 & 0.03 & 0.02 & 0.02 & 0.02 & 0.01 & \textbf{0.2} & 0.01 & 0.04 & 0.06 & 0.01 & 0.06 \\ 
 & 1 & 0.23 & 0.13 & 0.04 & 0.11 & 0.07 & 0.19 & 0.02 & 0.18 & 0.16 & 0.04 & \textbf{0.34} \\ 
 & 5 & 0.36 & 0.24 & 0.07 & 0.16 & 0.16 & 0.35 & 0.05 & 0.38 & 0.23 & 0.05 & \textbf{0.67} \\ 
 & 10 & 0.37 & 0.25 & 0.06 & 0.09 & 0.28 & 0.36 & 0.06 & 0.39 & 0.17 & 0.05 & \textbf{0.56} \\ 
\multirow{4}{3em}{synonym}  & 0 & 0.0 & 0.0 & 0.0 & 0.0 & 0.0 & \textbf{0.02} & 0.0 & 0.0 & 0.0 & 0.0 & 0.0 \\ 
 & 1 & 0.01 & 0.01 & 0.0 & 0.0 & 0.0 & \textbf{0.07} & 0.0 & 0.01 & 0.01 & 0.0 & 0.03 \\ 
 & 5 & 0.01 & 0.02 & 0.0 & 0.01 & 0.0 & 0.05 & 0.0 & 0.03 & 0.01 & 0.0 & \textbf{0.06} \\ 
 & 10 & 0.01 & 0.02 & 0.0 & 0.01 & 0.01 & \textbf{0.09} & 0.0 & 0.03 & 0.01 & 0.0 & 0.06 \\

    \bottomrule
  \end{tabular}
  \end{small}
  \caption{OPT-1.3B perturbation test on ICL tasks}
  \label{tab: perturbation appendix opt1.3B icl}
\end{table*}

\begin{table*}
  \centering
  \begin{small}
  \begin{tabular}{@{}c@{~~}c@{~~}c@{~~}c@{~~}c@{~~}c@{~~}c@{~~}c@{~~}c@{~~}c@{~~}c@{~~}c@{~~}c@{~~}}
    \toprule
     
    Task   & n-shots & Random & Index [$\leftarrow$] & Index & CM2 [$\leftarrow$] & CM2 & CM1 [$\leftarrow$] & CM1 & FA [$\leftarrow$] & FA & RA [$\leftarrow$] & RA \\
    \midrule

\multirow{4}{3em}{adjective-v-verb-3}  & 0 & 0.06 & 0.03 & 0.03 & 0.15 & 0.01 & \textbf{0.32} & 0.01 & 0.03 & 0.05 & 0.01 & 0.17 \\ 
 & 1 & 0.07 & 0.13 & 0.03 & 0.2 & 0.02 & \textbf{0.47} & 0.02 & 0.06 & 0.09 & 0.02 & 0.31 \\ 
 & 5 & 0.11 & 0.2 & 0.03 & 0.07 & 0.07 & \textbf{0.48} & 0.04 & 0.22 & 0.12 & 0.03 & 0.45 \\ 
 & 10 & 0.12 & 0.24 & 0.03 & 0.06 & 0.08 & 0.24 & 0.03 & 0.27 & 0.14 & 0.03 & \textbf{0.51} \\ 
\multirow{4}{3em}{alphabetic-ally-first-3}  & 0 & 0.2 & 0.04 & 0.04 & 0.25 & 0.01 & \textbf{0.32} & 0.01 & 0.04 & 0.05 & 0.02 & 0.18 \\ 
 & 1 & 0.11 & 0.12 & 0.02 & 0.09 & 0.08 & \textbf{0.24} & 0.02 & 0.07 & 0.07 & 0.02 & 0.23 \\ 
 & 5 & 0.17 & 0.14 & 0.02 & 0.05 & 0.07 & 0.1 & 0.04 & 0.14 & 0.1 & 0.03 & \textbf{0.25} \\ 
 & 10 & 0.16 & 0.14 & 0.02 & 0.05 & 0.08 & 0.22 & 0.02 & 0.16 & 0.08 & 0.03 & \textbf{0.25} \\ 
\multirow{4}{3em}{animal-v-object-3}  & 0 & 0.1 & 0.04 & 0.04 & 0.26 & 0.01 & \textbf{0.32} & 0.01 & 0.04 & 0.05 & 0.01 & 0.24 \\ 
 & 1 & 0.23 & 0.12 & 0.03 & 0.27 & 0.02 & \textbf{0.46} & 0.02 & 0.09 & 0.07 & 0.03 & 0.3 \\ 
 & 5 & 0.27 & 0.15 & 0.02 & 0.14 & 0.07 & 0.28 & 0.03 & 0.21 & 0.13 & 0.03 & \textbf{0.41} \\ 
 & 10 & 0.27 & 0.19 & 0.03 & 0.1 & 0.08 & 0.36 & 0.04 & 0.27 & 0.13 & 0.04 & \textbf{0.46} \\ 
\multirow{4}{3em}{antonym}  & 0 & 0.01 & 0.0 & 0.0 & 0.03 & 0.0 & \textbf{0.07} & 0.0 & 0.0 & 0.0 & 0.0 & 0.02 \\ 
 & 1 & 0.08 & 0.1 & 0.04 & 0.03 & 0.02 & \textbf{0.25} & 0.01 & 0.06 & 0.05 & 0.01 & 0.19 \\ 
 & 5 & 0.14 & 0.26 & 0.03 & 0.04 & 0.04 & 0.21 & 0.03 & 0.26 & 0.1 & 0.03 & \textbf{0.43} \\ 
 & 10 & 0.15 & 0.28 & 0.03 & 0.04 & 0.04 & 0.41 & 0.03 & 0.26 & 0.1 & 0.03 & \textbf{0.46} \\ 
\multirow{4}{3em}{capitalize}  & 0 & 0.0 & 0.0 & 0.0 & 0.13 & 0.0 & \textbf{0.17} & 0.0 & 0.0 & 0.0 & 0.0 & 0.0 \\ 
 & 1 & 0.07 & 0.23 & 0.04 & 0.03 & 0.02 & \textbf{0.34} & 0.02 & 0.08 & 0.07 & 0.02 & 0.28 \\ 
 & 5 & 0.2 & 0.46 & 0.06 & 0.08 & 0.3 & \textbf{0.77} & 0.05 & 0.45 & 0.19 & 0.05 & 0.68 \\ 
 & 10 & 0.21 & 0.46 & 0.05 & 0.07 & 0.29 & \textbf{0.74} & 0.05 & 0.47 & 0.18 & 0.05 & 0.7 \\ 
\multirow{4}{3em}{choose-middle-of-3}  & 0 & 0.01 & 0.01 & 0.0 & \textbf{0.14} & 0.0 & 0.12 & 0.0 & 0.01 & 0.0 & 0.0 & 0.02 \\ 
 & 1 & 0.09 & 0.16 & 0.02 & 0.03 & 0.07 & \textbf{0.53} & 0.02 & 0.05 & 0.06 & 0.02 & 0.29 \\ 
 & 5 & 0.13 & 0.25 & 0.03 & 0.04 & 0.12 & \textbf{0.38} & 0.03 & 0.21 & 0.12 & 0.04 & 0.36 \\ 
 & 10 & 0.15 & 0.31 & 0.04 & 0.05 & 0.11 & \textbf{0.51} & 0.04 & 0.32 & 0.13 & 0.06 & 0.47 \\ 
\multirow{4}{3em}{country-capital}  & 0 & 0.01 & 0.01 & 0.01 & \textbf{0.05} & 0.0 & \textbf{0.05} & 0.0 & 0.01 & 0.01 & 0.0 & 0.03 \\ 
 & 1 & 0.2 & 0.3 & 0.06 & 0.05 & 0.1 & 0.19 & 0.04 & 0.27 & 0.1 & 0.04 & \textbf{0.48} \\ 
 & 5 & 0.28 & 0.42 & 0.07 & 0.05 & 0.15 & 0.33 & 0.05 & 0.4 & 0.13 & 0.06 & \textbf{0.68} \\ 
 & 10 & 0.27 & 0.44 & 0.06 & 0.06 & 0.11 & 0.1 & 0.05 & 0.44 & 0.14 & 0.07 & \textbf{0.68} \\ 
\multirow{4}{3em}{english-french}  & 0 & 0.0 & 0.0 & 0.0 & 0.01 & 0.0 & \textbf{0.02} & 0.0 & 0.0 & 0.0 & 0.0 & 0.0 \\ 
 & 1 & 0.11 & 0.23 & 0.03 & 0.03 & 0.06 & 0.13 & 0.03 & 0.13 & 0.05 & 0.03 & \textbf{0.27} \\ 
 & 5 & 0.19 & 0.37 & 0.04 & 0.06 & 0.11 & 0.44 & 0.04 & 0.33 & 0.09 & 0.04 & \textbf{0.52} \\ 
 & 10 & 0.2 & 0.38 & 0.04 & 0.08 & 0.11 & 0.28 & 0.04 & 0.37 & 0.09 & 0.04 & \textbf{0.55} \\ 
\multirow{4}{3em}{lowercase-last-letter}  & 0 & 0.0 & 0.0 & 0.0 & 0.0 & 0.0 & 0.0 & 0.0 & 0.0 & 0.0 & 0.0 & 0.0 \\ 
 & 1 & 0.03 & 0.05 & 0.01 & 0.04 & 0.01 & \textbf{0.11} & 0.01 & 0.04 & 0.03 & 0.01 & 0.1 \\ 
 & 5 & 0.05 & 0.1 & 0.01 & 0.02 & 0.09 & 0.13 & 0.02 & 0.08 & 0.04 & 0.01 & \textbf{0.16} \\ 
 & 10 & 0.05 & 0.09 & 0.01 & 0.12 & 0.04 & 0.04 & 0.08 & 0.13 & 0.04 & 0.01 & \textbf{0.15} \\ 
\multirow{4}{3em}{next-item}  & 0 & 0.05 & 0.06 & 0.02 & 0.12 & 0.01 & \textbf{0.17} & 0.01 & 0.06 & 0.01 & 0.02 & 0.13 \\ 
 & 1 & 0.09 & 0.25 & 0.03 & 0.08 & 0.03 & \textbf{0.42} & 0.02 & 0.16 & 0.07 & 0.02 & 0.38 \\ 
 & 5 & 0.19 & 0.38 & 0.04 & 0.09 & 0.05 & 0.42 & 0.03 & 0.33 & 0.13 & 0.04 & \textbf{0.58} \\ 
 & 10 & 0.24 & 0.41 & 0.04 & 0.05 & 0.09 & 0.32 & 0.04 & 0.34 & 0.13 & 0.04 & \textbf{0.62} \\ 
\multirow{4}{3em}{person-sport}  & 0 & 0.0 & 0.0 & 0.0 & 0.0 & 0.0 & 0.0 & 0.0 & 0.0 & 0.0 & 0.0 & 0.0 \\ 
 & 1 & 0.18 & 0.38 & 0.05 & 0.11 & 0.19 & 0.34 & 0.04 & 0.21 & 0.11 & 0.05 & \textbf{0.59} \\ 
 & 5 & 0.27 & 0.47 & 0.07 & 0.08 & 0.2 & 0.39 & 0.05 & 0.46 & 0.13 & 0.07 & \textbf{0.66} \\ 
 & 10 & 0.28 & 0.5 & 0.08 & 0.16 & 0.2 & 0.37 & 0.08 & 0.53 & 0.12 & 0.07 & \textbf{0.62} \\ 
\multirow{4}{3em}{present-past}  & 0 & \textbf{0.01} & 0.0 & 0.0 & \textbf{0.01} & 0.0 & \textbf{0.01} & 0.0 & 0.0 & 0.0 & 0.0 & \textbf{0.01} \\ 
 & 1 & 0.04 & 0.16 & 0.06 & 0.04 & 0.03 & \textbf{0.55} & 0.02 & 0.13 & 0.09 & 0.02 & 0.29 \\ 
 & 5 & 0.17 & 0.48 & 0.08 & 0.06 & 0.3 & \textbf{0.8} & 0.05 & 0.42 & 0.22 & 0.05 & 0.76 \\ 
 & 10 & 0.26 & 0.5 & 0.06 & 0.06 & 0.34 & \textbf{0.82} & 0.05 & 0.51 & 0.22 & 0.05 & 0.69 \\ 
\multirow{4}{3em}{prev-item}  & 0 & 0.03 & 0.04 & 0.01 & 0.08 & 0.01 & \textbf{0.11} & 0.01 & 0.04 & 0.01 & 0.01 & 0.07 \\ 
 & 1 & 0.04 & 0.14 & 0.02 & 0.05 & 0.02 & 0.1 & 0.01 & 0.09 & 0.04 & 0.01 & \textbf{0.21} \\ 
 & 5 & 0.06 & 0.25 & 0.03 & 0.06 & 0.04 & 0.2 & 0.02 & 0.22 & 0.08 & 0.03 & \textbf{0.41} \\ 
 & 10 & 0.07 & 0.3 & 0.03 & 0.04 & 0.08 & 0.14 & 0.04 & 0.29 & 0.08 & 0.03 & \textbf{0.45} \\ 
\multirow{4}{3em}{singular-plural}  & 0 & 0.11 & 0.05 & 0.02 & 0.17 & 0.01 & \textbf{0.23} & 0.01 & 0.05 & 0.03 & 0.04 & 0.12 \\ 
 & 1 & 0.2 & 0.31 & 0.07 & 0.12 & 0.08 & 0.48 & 0.04 & 0.23 & 0.15 & 0.04 & \textbf{0.52} \\ 
 & 5 & 0.33 & 0.45 & 0.07 & 0.1 & 0.31 & \textbf{0.76} & 0.06 & 0.48 & 0.21 & 0.05 & \textbf{0.76} \\ 
 & 10 & 0.37 & 0.49 & 0.06 & 0.09 & 0.27 & 0.72 & 0.05 & 0.49 & 0.19 & 0.05 & \textbf{0.78} \\ 
\multirow{4}{3em}{synonym}  & 0 & 0.0 & 0.0 & 0.0 & 0.0 & 0.0 & 0.0 & 0.0 & 0.0 & 0.0 & 0.0 & 0.0 \\ 
 & 1 & 0.01 & 0.03 & 0.01 & 0.01 & 0.0 & 0.04 & 0.0 & 0.02 & 0.01 & 0.0 & \textbf{0.05} \\ 
 & 5 & 0.02 & 0.06 & 0.01 & 0.01 & 0.02 & 0.06 & 0.01 & 0.08 & 0.02 & 0.01 & \textbf{0.11} \\ 
 & 10 & 0.02 & 0.06 & 0.01 & 0.01 & 0.03 & 0.02 & 0.01 & 0.07 & 0.03 & 0.01 & \textbf{0.11} \\ 

    \bottomrule
  \end{tabular}
  \end{small}
  \caption{GPT-j perturbation test on ICL tasks}
  \label{tab: perturbation appendix gptj icl}
\end{table*}

\begin{table*}
  \centering
  \begin{small}
  \begin{tabular}{@{}c@{~~}c@{~~}c@{~~}c@{~~}c@{~~}c@{~~}c@{~~}c@{~~}c@{~~}c@{~~}c@{~~}c@{~~}c@{~~}}
    \toprule
     
    Task   & n-shots & Random & Index [$\leftarrow$] & Index & CM2 [$\leftarrow$] & CM2 & CM1 [$\leftarrow$] & CM1 & FA [$\leftarrow$] & FA & RA [$\leftarrow$] & RA \\
    \midrule

\multirow{4}{3em}{adjective-v-verb-3}  & 0 & 0.1 & 0.04 & 0.03 & 0.17 & 0.01 & \textbf{0.35} & 0.01 & 0.14 & 0.04 & 0.01 & 0.15 \\ 
 & 1 & 0.16 & 0.2 & 0.04 & 0.17 & 0.03 & \textbf{0.4} & 0.03 & 0.28 & 0.05 & 0.03 & 0.37 \\ 
 & 5 & 0.25 & 0.21 & 0.04 & 0.08 & 0.06 & 0.11 & 0.03 & 0.35 & 0.07 & 0.03 & \textbf{0.45} \\ 
 & 10 & 0.29 & 0.26 & 0.04 & 0.09 & 0.09 & 0.12 & 0.04 & 0.36 & 0.05 & 0.04 & \textbf{0.48} \\ 
\multirow{4}{3em}{alphabetic-ally-first-3}  & 0 & 0.03 & 0.04 & 0.03 & 0.14 & 0.01 & \textbf{0.17} & 0.01 & 0.12 & 0.04 & 0.01 & 0.11 \\ 
 & 1 & 0.08 & 0.14 & 0.03 & 0.1 & 0.06 & 0.13 & 0.02 & 0.19 & 0.04 & 0.02 & \textbf{0.22} \\ 
 & 5 & 0.08 & 0.14 & 0.03 & 0.08 & 0.04 & 0.08 & 0.02 & 0.19 & 0.04 & 0.02 & \textbf{0.24} \\ 
 & 10 & 0.07 & 0.13 & 0.02 & 0.06 & 0.06 & 0.07 & 0.02 & 0.19 & 0.02 & 0.02 & \textbf{0.26} \\ 
\multirow{4}{3em}{animal-v-object-3}  & 0 & 0.08 & 0.07 & 0.03 & 0.12 & 0.02 & 0.09 & 0.01 & \textbf{0.15} & 0.04 & 0.02 & \textbf{0.15} \\ 
 & 1 & 0.12 & 0.13 & 0.03 & 0.16 & 0.05 & 0.17 & 0.03 & 0.22 & 0.05 & 0.02 & \textbf{0.27} \\ 
 & 5 & 0.22 & 0.18 & 0.03 & 0.09 & 0.04 & 0.11 & 0.03 & 0.28 & 0.06 & 0.03 & \textbf{0.39} \\ 
 & 10 & 0.27 & 0.22 & 0.04 & 0.08 & 0.06 & 0.04 & 0.06 & 0.34 & 0.05 & 0.03 & \textbf{0.43} \\ 
\multirow{4}{3em}{antonym}  & 0 & 0.0 & 0.0 & 0.0 & 0.0 & 0.0 & 0.0 & 0.0 & 0.0 & 0.0 & 0.0 & \textbf{0.01} \\ 
 & 1 & 0.07 & 0.1 & 0.02 & 0.06 & 0.02 & 0.12 & 0.02 & 0.15 & 0.02 & 0.02 & \textbf{0.21} \\ 
 & 5 & 0.17 & 0.25 & 0.04 & 0.07 & 0.05 & 0.08 & 0.03 & 0.29 & 0.04 & 0.03 & \textbf{0.34} \\ 
 & 10 & 0.19 & 0.28 & 0.04 & 0.05 & 0.07 & 0.1 & 0.03 & 0.19 & 0.04 & 0.03 & \textbf{0.33} \\ 
\multirow{4}{3em}{capitalize}  & 0 & 0.05 & 0.08 & 0.02 & 0.08 & 0.02 & \textbf{0.52} & 0.02 & 0.14 & 0.02 & 0.02 & 0.16 \\ 
 & 1 & 0.19 & 0.36 & 0.04 & 0.07 & 0.03 & 0.07 & 0.03 & 0.41 & 0.05 & 0.03 & \textbf{0.58} \\ 
 & 5 & 0.35 & 0.45 & 0.05 & 0.07 & 0.15 & 0.14 & 0.05 & 0.65 & 0.06 & 0.05 & \textbf{0.76} \\ 
 & 10 & 0.36 & 0.46 & 0.05 & 0.1 & 0.14 & 0.33 & 0.05 & 0.61 & 0.05 & 0.05 & \textbf{0.76} \\ 
\multirow{4}{3em}{choose-middle-of-3}  & 0 & 0.01 & 0.01 & 0.0 & 0.05 & 0.0 & \textbf{0.16} & 0.0 & 0.02 & 0.0 & 0.01 & 0.02 \\ 
 & 1 & 0.09 & 0.19 & 0.03 & 0.05 & 0.1 & \textbf{0.38} & 0.03 & 0.22 & 0.04 & 0.03 & 0.3 \\ 
 & 5 & 0.16 & 0.17 & 0.03 & 0.04 & 0.03 & 0.22 & 0.02 & 0.21 & 0.05 & 0.03 & \textbf{0.36} \\ 
 & 10 & 0.14 & 0.2 & 0.03 & 0.07 & 0.05 & 0.06 & 0.03 & 0.24 & 0.04 & 0.03 & \textbf{0.43} \\ 
\multirow{4}{3em}{country-capital}  & 0 & 0.02 & 0.0 & 0.0 & 0.0 & 0.02 & 0.0 & 0.0 & \textbf{0.04} & 0.01 & 0.0 & 0.02 \\ 
 & 1 & 0.17 & 0.39 & 0.04 & 0.21 & 0.09 & 0.17 & 0.04 & 0.5 & 0.05 & 0.05 & \textbf{0.52} \\ 
 & 5 & 0.22 & 0.43 & 0.05 & 0.15 & 0.08 & 0.15 & 0.05 & \textbf{0.58} & 0.07 & 0.05 & 0.56 \\ 
 & 10 & 0.24 & 0.42 & 0.05 & 0.12 & 0.22 & 0.14 & 0.05 & 0.48 & 0.05 & 0.05 & \textbf{0.57} \\ 
\multirow{4}{3em}{english-french}  & 0 & \textbf{0.02} & 0.0 & 0.0 & \textbf{0.02} & 0.0 & \textbf{0.02} & 0.0 & 0.01 & 0.0 & 0.0 & 0.01 \\ 
 & 1 & 0.08 & 0.24 & 0.03 & 0.05 & 0.07 & 0.06 & 0.03 & \textbf{0.31} & 0.03 & 0.03 & \textbf{0.31} \\ 
 & 5 & 0.13 & 0.36 & 0.04 & 0.07 & 0.1 & 0.14 & 0.04 & 0.44 & 0.04 & 0.04 & \textbf{0.47} \\ 
 & 10 & 0.14 & 0.37 & 0.04 & 0.07 & 0.06 & 0.17 & 0.04 & 0.38 & 0.04 & 0.04 & \textbf{0.45} \\ 
\multirow{4}{3em}{lowercase-last-letter}  & 0 & 0.0 & 0.0 & 0.0 & 0.0 & 0.0 & 0.0 & 0.0 & 0.0 & 0.0 & 0.0 & 0.0 \\ 
 & 1 & 0.02 & 0.06 & 0.01 & 0.01 & 0.01 & 0.03 & 0.01 & \textbf{0.09} & 0.01 & 0.01 & \textbf{0.09} \\ 
 & 5 & 0.05 & 0.11 & 0.02 & 0.02 & 0.06 & 0.02 & 0.02 & 0.15 & 0.02 & 0.01 & \textbf{0.16} \\ 
 & 10 & 0.05 & 0.11 & 0.02 & 0.07 & 0.03 & 0.09 & 0.02 & 0.12 & 0.02 & 0.01 & \textbf{0.16} \\ 
\multirow{4}{3em}{next-item}  & 0 & 0.11 & 0.08 & 0.06 & 0.11 & 0.05 & 0.14 & 0.01 & 0.1 & 0.07 & 0.04 & \textbf{0.15} \\ 
 & 1 & 0.1 & 0.19 & 0.06 & 0.12 & 0.07 & 0.12 & 0.07 & 0.26 & 0.07 & 0.05 & \textbf{0.35} \\ 
 & 5 & 0.19 & 0.39 & 0.08 & 0.13 & 0.1 & 0.18 & 0.05 & 0.47 & 0.09 & 0.09 & \textbf{0.66} \\ 
 & 10 & 0.2 & 0.44 & 0.07 & 0.09 & 0.07 & 0.23 & 0.07 & 0.39 & 0.06 & 0.08 & \textbf{0.62} \\ 
\multirow{4}{3em}{person-sport}  & 0 & 0.0 & 0.0 & 0.0 & 0.0 & 0.0 & 0.0 & 0.0 & 0.0 & 0.0 & 0.0 & 0.0 \\ 
 & 1 & 0.18 & 0.48 & 0.05 & 0.13 & 0.07 & 0.14 & 0.04 & 0.48 & 0.06 & 0.05 & \textbf{0.58} \\ 
 & 5 & 0.25 & 0.52 & 0.07 & 0.06 & 0.16 & 0.55 & 0.06 & 0.55 & 0.08 & 0.05 & \textbf{0.64} \\ 
 & 10 & 0.28 & 0.53 & 0.07 & 0.07 & 0.12 & 0.19 & 0.05 & 0.53 & 0.07 & 0.05 & \textbf{0.63} \\ 
\multirow{4}{3em}{present-past}  & 0 & 0.02 & 0.02 & 0.01 & \textbf{0.06} & 0.0 & 0.0 & 0.0 & 0.05 & 0.01 & 0.01 & 0.04 \\ 
 & 1 & 0.14 & 0.36 & 0.05 & 0.06 & 0.11 & 0.18 & 0.04 & 0.46 & 0.05 & 0.04 & \textbf{0.56} \\ 
 & 5 & 0.26 & 0.44 & 0.05 & 0.06 & 0.14 & 0.34 & 0.05 & 0.55 & 0.06 & 0.05 & \textbf{0.7} \\ 
 & 10 & 0.19 & 0.44 & 0.05 & 0.07 & 0.14 & 0.36 & 0.05 & 0.6 & 0.05 & 0.05 & \textbf{0.75} \\ 
\multirow{4}{3em}{prev-item}  & 0 & 0.08 & 0.08 & 0.05 & \textbf{0.12} & 0.05 & \textbf{0.12} & 0.01 & 0.1 & 0.04 & 0.04 & 0.11 \\ 
 & 1 & 0.07 & 0.16 & 0.06 & 0.07 & 0.07 & 0.05 & 0.06 & 0.16 & 0.05 & 0.06 & \textbf{0.18} \\ 
 & 5 & 0.08 & 0.29 & 0.04 & 0.06 & 0.07 & 0.19 & 0.05 & 0.29 & 0.06 & 0.07 & \textbf{0.44} \\ 
 & 10 & 0.1 & 0.35 & 0.06 & 0.12 & 0.07 & 0.25 & 0.06 & 0.32 & 0.06 & 0.08 & \textbf{0.53} \\ 
\multirow{4}{3em}{singular-plural}  & 0 & \textbf{0.09} & 0.0 & 0.01 & 0.06 & 0.01 & 0.0 & 0.0 & 0.08 & 0.05 & 0.01 & 0.05 \\ 
 & 1 & 0.19 & 0.43 & 0.06 & 0.29 & 0.06 & 0.19 & 0.05 & 0.53 & 0.09 & 0.05 & \textbf{0.68} \\ 
 & 5 & 0.24 & 0.45 & 0.05 & 0.12 & 0.14 & 0.37 & 0.05 & 0.64 & 0.09 & 0.05 & \textbf{0.76} \\ 
 & 10 & 0.23 & 0.45 & 0.05 & 0.12 & 0.11 & 0.44 & 0.05 & 0.61 & 0.05 & 0.05 & \textbf{0.75} \\ 
\multirow{4}{3em}{synonym}  & 0 & 0.0 & 0.0 & 0.0 & \textbf{0.01} & 0.0 & 0.0 & 0.0 & \textbf{0.01} & 0.0 & 0.0 & 0.0 \\ 
 & 1 & 0.03 & 0.06 & 0.02 & 0.03 & 0.02 & 0.04 & 0.01 & 0.12 & 0.02 & 0.01 & \textbf{0.13} \\ 
 & 5 & 0.08 & 0.12 & 0.02 & 0.03 & 0.04 & 0.03 & 0.02 & 0.19 & 0.03 & 0.02 & \textbf{0.25} \\ 
 & 10 & 0.09 & 0.12 & 0.02 & 0.03 & 0.04 & 0.04 & 0.02 & 0.17 & 0.02 & 0.02 & \textbf{0.25} \\ 

    \bottomrule
  \end{tabular}
  \end{small}
  \caption{Llama2-7B perturbation test on ICL tasks}
  \label{tab: perturbation appendix llama2-7B icl}
\end{table*}

\begin{table*}
  \centering
  \begin{small}
  \begin{tabular}{@{}c@{~~}c@{~~}c@{~~}c@{~~}c@{~~}c@{~~}c@{~~}c@{~~}c@{~~}c@{~~}c@{~~}c@{~~}c@{~~}}
    \toprule

    Task   & n-shots & Random & Index [$\leftarrow$] & Index & CM2 [$\leftarrow$] & CM2 & CM1 [$\leftarrow$] & CM1 & FA [$\leftarrow$] & FA & RA [$\leftarrow$] & RA \\
    \midrule

\multirow{4}{3em}{company-hq}  & 0 & 0.15 & 0.09 & 0.05 & 0.13 & 0.09 & \textbf{0.19} & 0.05 & 0.15 & 0.06 & 0.04 & 0.18 \\ 
 & 1 & 0.07 & 0.09 & 0.04 & 0.12 & 0.04 & \textbf{0.19} & 0.03 & 0.12 & 0.05 & 0.02 & 0.16 \\ 
 & 5 & 0.14 & 0.14 & 0.05 & 0.11 & 0.04 & \textbf{0.21} & 0.03 & 0.14 & 0.07 & 0.03 & 0.2 \\ 
 & 10 & 0.14 & 0.13 & 0.05 & 0.12 & 0.03 & 0.19 & 0.05 & 0.13 & 0.08 & 0.04 & \textbf{0.2} \\ 
\multirow{4}{3em}{country-capital}  & 0 & 0.02 & 0.03 & 0.05 & 0.36 & 0.01 & \textbf{0.61} & 0.01 & 0.08 & 0.06 & 0.01 & 0.07 \\ 
 & 1 & 0.27 & 0.31 & 0.13 & 0.21 & 0.13 & \textbf{0.58} & 0.04 & 0.32 & 0.14 & 0.04 & 0.57 \\ 
 & 5 & 0.43 & 0.34 & 0.12 & 0.19 & 0.15 & 0.52 & 0.1 & 0.32 & 0.13 & 0.07 & \textbf{0.57} \\ 
 & 10 & 0.41 & 0.33 & 0.12 & 0.24 & 0.1 & 0.36 & 0.07 & 0.28 & 0.17 & 0.07 & \textbf{0.51} \\ 
\multirow{4}{3em}{landmark-in-country}  & 0 & 0.06 & 0.05 & 0.03 & 0.08 & 0.01 & \textbf{0.35} & 0.01 & 0.12 & 0.05 & 0.01 & 0.18 \\ 
 & 1 & 0.14 & 0.14 & 0.06 & 0.11 & 0.04 & \textbf{0.29} & 0.04 & 0.11 & 0.08 & 0.03 & 0.28 \\ 
 & 5 & 0.18 & 0.17 & 0.07 & 0.07 & 0.07 & 0.23 & 0.02 & 0.2 & 0.09 & 0.04 & \textbf{0.33} \\ 
 & 10 & 0.2 & 0.18 & 0.07 & 0.1 & 0.08 & 0.2 & 0.02 & 0.2 & 0.1 & 0.03 & \textbf{0.33} \\ 
\multirow{4}{3em}{person-plays-pro-sport}  & 0 & 0.32 & 0.24 & 0.08 & 0.36 & 0.05 & \textbf{0.59} & 0.03 & 0.48 & 0.09 & 0.07 & 0.57 \\ 
 & 1 & 0.32 & 0.27 & 0.1 & 0.24 & 0.25 & 0.42 & 0.07 & 0.32 & 0.12 & 0.1 & \textbf{0.49} \\ 
 & 5 & 0.3 & 0.25 & 0.1 & 0.3 & 0.15 & 0.41 & 0.13 & 0.33 & 0.12 & 0.1 & \textbf{0.46} \\ 
 & 10 & 0.28 & 0.31 & 0.12 & 0.27 & 0.16 & 0.36 & 0.04 & 0.33 & 0.13 & 0.09 & \textbf{0.49} \\ 
\multirow{4}{3em}{present-past}  & 0 & 0.0 & 0.0 & 0.0 & 0.01 & 0.0 & \textbf{0.04} & 0.0 & 0.0 & 0.01 & 0.0 & 0.0 \\ 
 & 1 & 0.18 & 0.38 & 0.13 & 0.06 & 0.08 & \textbf{0.58} & 0.04 & 0.1 & 0.15 & 0.04 & 0.35 \\ 
 & 5 & 0.38 & 0.46 & 0.15 & 0.12 & 0.23 & 0.56 & 0.05 & 0.34 & 0.23 & 0.05 & \textbf{0.61} \\ 
 & 10 & 0.42 & 0.46 & 0.14 & 0.1 & 0.27 & \textbf{0.67} & 0.08 & 0.41 & 0.22 & 0.08 & 0.63 \\ 
\multirow{4}{3em}{product-by-company}  & 0 & 0.19 & 0.14 & 0.06 & 0.2 & 0.08 & \textbf{0.46} & 0.02 & 0.24 & 0.08 & 0.02 & 0.31 \\ 
 & 1 & 0.25 & 0.25 & 0.12 & 0.33 & 0.09 & 0.34 & 0.03 & 0.24 & 0.18 & 0.07 & \textbf{0.44} \\ 
 & 5 & 0.33 & 0.29 & 0.17 & 0.32 & 0.18 & 0.49 & 0.07 & 0.33 & 0.21 & 0.1 & \textbf{0.55} \\ 
 & 10 & 0.32 & 0.29 & 0.15 & 0.36 & 0.15 & 0.46 & 0.08 & 0.35 & 0.2 & 0.08 & \textbf{0.53} \\
 
    \bottomrule
  \end{tabular}
  \end{small}
  \caption{GPT2-xl perturbation test on tasks of natural questions. We also provide results where we prompt the models with ICL-like prompting, where a few labeled example proceed the actual prompt the model answer. Hence, the 0-shot is the basic natural question task, and the other n-shots are provided for comprehensive examination.}
  \label{tab: perturbation appendix gpt2xl natural}
\end{table*}

\begin{table*}
  \centering
  \begin{small}
  \begin{tabular}{@{}c@{~~}c@{~~}c@{~~}c@{~~}c@{~~}c@{~~}c@{~~}c@{~~}c@{~~}c@{~~}c@{~~}c@{~~}c@{~~}}
    \toprule

    Task   & n-shots & Random & Index [$\leftarrow$] & Index & CM2 [$\leftarrow$] & CM2 & CM1 [$\leftarrow$] & CM1 & FA [$\leftarrow$] & FA & RA [$\leftarrow$] & RA \\
    \midrule

\multirow{4}{3em}{company-hq}  & 0 & 0.16 & 0.05 & 0.04 & 0.18 & 0.02 & \textbf{0.27} & 0.02 & 0.2 & 0.05 & 0.03 & 0.23 \\ 
 & 1 & 0.11 & 0.08 & 0.04 & 0.19 & 0.04 & \textbf{0.27} & 0.02 & 0.17 & 0.07 & 0.02 & 0.23 \\ 
 & 5 & 0.15 & 0.12 & 0.04 & 0.16 & 0.02 & 0.21 & 0.04 & 0.2 & 0.07 & 0.03 & \textbf{0.25} \\ 
 & 10 & 0.16 & 0.14 & 0.03 & 0.08 & 0.06 & 0.21 & 0.04 & 0.2 & 0.08 & 0.04 & \textbf{0.25} \\ 
\multirow{4}{3em}{country-capital}  & 0 & 0.04 & 0.01 & 0.03 & 0.03 & 0.01 & \textbf{0.56} & 0.01 & 0.15 & 0.04 & 0.01 & 0.2 \\ 
 & 1 & 0.33 & 0.16 & 0.12 & 0.1 & 0.34 & \textbf{0.67} & 0.05 & 0.34 & 0.22 & 0.05 & 0.62 \\ 
 & 5 & 0.34 & 0.17 & 0.12 & 0.15 & 0.3 & 0.44 & 0.12 & 0.31 & 0.22 & 0.05 & \textbf{0.56} \\ 
 & 10 & 0.31 & 0.17 & 0.11 & 0.11 & 0.28 & 0.26 & 0.17 & 0.29 & 0.19 & 0.05 & \textbf{0.5} \\ 
\multirow{4}{3em}{landmark-in-country}  & 0 & 0.2 & 0.07 & 0.04 & 0.22 & 0.02 & \textbf{0.32} & 0.02 & 0.26 & 0.05 & 0.02 & 0.29 \\ 
 & 1 & 0.2 & 0.11 & 0.06 & 0.17 & 0.06 & 0.21 & 0.05 & 0.21 & 0.1 & 0.04 & \textbf{0.37} \\ 
 & 5 & 0.21 & 0.11 & 0.06 & 0.17 & 0.06 & 0.26 & 0.04 & 0.28 & 0.11 & 0.04 & \textbf{0.37} \\ 
 & 10 & 0.21 & 0.11 & 0.04 & 0.1 & 0.07 & 0.19 & 0.05 & 0.26 & 0.1 & 0.03 & \textbf{0.35} \\ 
\multirow{4}{3em}{person-plays-pro-sport}  & 0 & 0.33 & 0.22 & 0.08 & 0.32 & 0.04 & 0.47 & 0.04 & 0.44 & 0.1 & 0.07 & \textbf{0.51} \\ 
 & 1 & 0.3 & 0.23 & 0.11 & 0.22 & 0.24 & 0.42 & 0.15 & 0.34 & 0.17 & 0.11 & \textbf{0.45} \\ 
 & 5 & 0.34 & 0.26 & 0.08 & 0.32 & 0.13 & 0.39 & 0.09 & 0.39 & 0.13 & 0.08 & \textbf{0.44} \\ 
 & 10 & 0.35 & 0.25 & 0.07 & 0.29 & 0.15 & \textbf{0.47} & 0.12 & 0.38 & 0.13 & 0.09 & 0.46 \\ 
\multirow{4}{3em}{present-past}  & 0 & 0.0 & 0.0 & 0.0 & 0.0 & 0.0 & \textbf{0.02} & 0.0 & 0.0 & 0.0 & 0.0 & 0.0 \\ 
 & 1 & 0.1 & 0.15 & 0.06 & 0.02 & 0.05 & \textbf{0.64} & 0.02 & 0.1 & 0.1 & 0.02 & 0.27 \\ 
 & 5 & 0.28 & 0.23 & 0.12 & 0.06 & 0.19 & \textbf{0.62} & 0.04 & 0.4 & 0.26 & 0.04 & 0.57 \\ 
 & 10 & 0.3 & 0.24 & 0.13 & 0.06 & 0.23 & 0.44 & 0.05 & 0.4 & 0.22 & 0.05 & \textbf{0.61} \\ 
\multirow{4}{3em}{product-by-company}  & 0 & 0.26 & 0.1 & 0.09 & 0.09 & 0.3 & \textbf{0.46} & 0.02 & 0.3 & 0.1 & 0.03 & 0.41 \\ 
 & 1 & 0.32 & 0.17 & 0.08 & 0.22 & 0.11 & \textbf{0.52} & 0.06 & 0.32 & 0.16 & 0.1 & 0.46 \\ 
 & 5 & 0.31 & 0.19 & 0.08 & 0.21 & 0.18 & 0.48 & 0.08 & 0.41 & 0.16 & 0.08 & \textbf{0.53} \\ 
 & 10 & 0.29 & 0.2 & 0.08 & 0.23 & 0.16 & 0.29 & 0.11 & 0.38 & 0.14 & 0.07 & \textbf{0.48} \\
 
    \bottomrule
  \end{tabular}
  \end{small}
  \caption{OPT-1.3B perturbation test on tasks of natural questions.}
  \label{tab: perturbation appendix opt1.3 natural}
\end{table*}

\begin{table*}
  \centering
  \begin{small}
  \begin{tabular}{@{}c@{~~}c@{~~}c@{~~}c@{~~}c@{~~}c@{~~}c@{~~}c@{~~}c@{~~}c@{~~}c@{~~}c@{~~}c@{~~}}
    \toprule

    Task   & n-shots & Random & Index [$\leftarrow$] & Index & CM2 [$\leftarrow$] & CM2 & CM1 [$\leftarrow$] & CM1 & FA [$\leftarrow$] & FA & RA [$\leftarrow$] & RA \\
    \midrule

    \multirow{4}{3em}{company-hq}  & 0 & 0.15 & 0.13 & 0.05 & 0.12 & 0.02 & 0.25 & 0.02 & 0.24 & 0.04 & 0.03 & \textbf{0.3} \\ 
 & 1 & 0.11 & 0.13 & 0.04 & 0.16 & 0.02 & \textbf{0.26} & 0.02 & 0.2 & 0.03 & 0.02 & 0.2 \\ 
 & 5 & 0.18 & 0.23 & 0.05 & 0.15 & 0.05 & 0.23 & 0.03 & 0.32 & 0.06 & 0.04 & \textbf{0.38} \\ 
 & 10 & 0.21 & 0.25 & 0.05 & 0.13 & 0.05 & 0.23 & 0.03 & 0.33 & 0.06 & 0.04 & \textbf{0.36} \\ 
\multirow{4}{3em}{country-capital}  & 0 & 0.31 & 0.28 & 0.07 & 0.48 & 0.04 & \textbf{0.72} & 0.04 & 0.48 & 0.13 & 0.05 & 0.53 \\ 
 & 1 & 0.44 & 0.44 & 0.1 & 0.35 & 0.19 & \textbf{0.67} & 0.05 & 0.53 & 0.16 & 0.12 & \textbf{0.67} \\ 
 & 5 & 0.43 & 0.43 & 0.09 & 0.13 & 0.23 & 0.31 & 0.06 & 0.51 & 0.18 & 0.14 & \textbf{0.67} \\ 
 & 10 & 0.43 & 0.44 & 0.09 & 0.13 & 0.25 & 0.47 & 0.07 & 0.47 & 0.16 & 0.14 & \textbf{0.68} \\ 
\multirow{4}{3em}{landmark-in-country}  & 0 & 0.21 & 0.19 & 0.02 & 0.28 & 0.02 & \textbf{0.43} & 0.02 & 0.26 & 0.02 & 0.02 & 0.35 \\ 
 & 1 & 0.23 & 0.33 & 0.04 & 0.19 & 0.06 & \textbf{0.51} & 0.03 & 0.42 & 0.06 & 0.04 & 0.45 \\ 
 & 5 & 0.26 & 0.35 & 0.05 & 0.12 & 0.06 & 0.5 & 0.04 & 0.45 & 0.09 & 0.04 & \textbf{0.51} \\ 
 & 10 & 0.26 & 0.37 & 0.05 & 0.21 & 0.06 & 0.5 & 0.04 & 0.49 & 0.09 & 0.04 & \textbf{0.53} \\ 
\multirow{4}{3em}{person-plays-pro-sport}  & 0 & 0.22 & 0.25 & 0.03 & 0.27 & 0.03 & \textbf{0.54} & 0.03 & 0.39 & 0.03 & 0.03 & 0.43 \\ 
 & 1 & 0.31 & 0.36 & 0.08 & 0.39 & 0.09 & 0.37 & 0.07 & \textbf{0.49} & 0.09 & 0.08 & \textbf{0.49} \\ 
 & 5 & 0.32 & 0.41 & 0.09 & 0.25 & 0.18 & \textbf{0.63} & 0.05 & 0.52 & 0.12 & 0.1 & 0.57 \\ 
 & 10 & 0.34 & 0.45 & 0.08 & 0.35 & 0.22 & 0.56 & 0.07 & \textbf{0.62} & 0.14 & 0.1 & 0.61 \\ 
\multirow{4}{3em}{present-past}  & 0 & 0.06 & 0.23 & 0.03 & 0.12 & 0.04 & \textbf{0.61} & 0.03 & 0.3 & 0.03 & 0.03 & 0.35 \\ 
 & 1 & 0.27 & 0.41 & 0.05 & 0.1 & 0.15 & \textbf{0.74} & 0.05 & 0.34 & 0.18 & 0.05 & 0.56 \\ 
 & 5 & 0.41 & 0.46 & 0.05 & 0.1 & 0.39 & \textbf{0.69} & 0.05 & 0.48 & 0.23 & 0.05 & 0.61 \\ 
 & 10 & 0.42 & 0.47 & 0.05 & 0.07 & 0.39 & 0.65 & 0.05 & 0.51 & 0.15 & 0.05 & \textbf{0.67} \\ 
\multirow{4}{3em}{product-by-company}  & 0 & 0.29 & 0.23 & 0.1 & \textbf{0.43} & 0.02 & 0.42 & 0.02 & 0.37 & 0.1 & 0.05 & 0.42 \\ 
 & 1 & 0.3 & 0.33 & 0.12 & 0.35 & 0.13 & 0.46 & 0.04 & 0.44 & 0.1 & 0.09 & \textbf{0.5} \\ 
 & 5 & 0.34 & 0.36 & 0.15 & 0.38 & 0.12 & 0.47 & 0.08 & 0.49 & 0.15 & 0.1 & \textbf{0.56} \\ 
 & 10 & 0.35 & 0.38 & 0.15 & 0.41 & 0.12 & \textbf{0.55} & 0.08 & 0.49 & 0.17 & 0.08 & 0.54 \\ 
 
    \bottomrule
  \end{tabular}
  \end{small}
  \caption{GPT-j perturbation test on tasks of natural questions.}
  \label{tab: perturbation appendix gptj natural}
\end{table*}

\begin{table*}
  \centering
  \begin{small}
  \begin{tabular}{@{}c@{~~}c@{~~}c@{~~}c@{~~}c@{~~}c@{~~}c@{~~}c@{~~}c@{~~}c@{~~}c@{~~}c@{~~}c@{~~}}
    \toprule

    Task   & n-shots & Random & Index [$\leftarrow$] & Index & CM2 [$\leftarrow$] & CM2 & CM1 [$\leftarrow$] & CM1 & FA [$\leftarrow$] & FA & RA [$\leftarrow$] & RA \\
    \midrule

\multirow{4}{3em}{company-hq}  & 0 & 0.17 & 0.23 & 0.04 & 0.2 & 0.08 & 0.03 & 0.03 & 0.3 & 0.05 & 0.03 & \textbf{0.32} \\ 
 & 1 & 0.11 & 0.26 & 0.04 & 0.19 & 0.05 & 0.24 & 0.03 & 0.28 & 0.04 & 0.03 & \textbf{0.3} \\ 
 & 5 & 0.2 & 0.3 & 0.05 & 0.24 & 0.04 & 0.12 & 0.03 & 0.31 & 0.04 & 0.04 & \textbf{0.37} \\ 
 & 10 & 0.19 & 0.32 & 0.05 & 0.26 & 0.05 & 0.11 & 0.04 & 0.3 & 0.04 & 0.04 & \textbf{0.39} \\ 
\multirow{4}{3em}{country-capital}  & 0 & 0.31 & 0.23 & 0.05 & 0.22 & 0.05 & 0.05 & 0.05 & \textbf{0.58} & 0.07 & 0.08 & 0.43 \\ 
 & 1 & 0.39 & 0.42 & 0.08 & 0.34 & 0.1 & \textbf{0.71} & 0.05 & 0.62 & 0.11 & 0.08 & 0.52 \\ 
 & 5 & 0.41 & 0.43 & 0.07 & 0.32 & 0.1 & 0.33 & 0.06 & \textbf{0.61} & 0.11 & 0.07 & 0.57 \\ 
 & 10 & 0.41 & 0.43 & 0.06 & 0.28 & 0.23 & 0.34 & 0.05 & 0.53 & 0.09 & 0.07 & \textbf{0.57} \\ 
\multirow{4}{3em}{landmark-in-country}  & 0 & 0.12 & 0.33 & 0.04 & 0.19 & 0.05 & 0.44 & 0.04 & 0.48 & 0.06 & 0.05 & \textbf{0.51} \\ 
 & 1 & 0.13 & 0.38 & 0.05 & 0.19 & 0.06 & 0.33 & 0.04 & 0.48 & 0.07 & 0.04 & \textbf{0.54} \\ 
 & 5 & 0.15 & 0.42 & 0.06 & 0.24 & 0.06 & 0.12 & 0.05 & 0.5 & 0.08 & 0.05 & \textbf{0.6} \\ 
 & 10 & 0.16 & 0.44 & 0.06 & 0.22 & 0.09 & 0.34 & 0.05 & 0.43 & 0.06 & 0.05 & \textbf{0.57} \\ 
\multirow{4}{3em}{person-plays-pro-sport}  & 0 & 0.21 & 0.33 & 0.04 & 0.27 & 0.04 & 0.39 & 0.04 & 0.43 & 0.04 & 0.04 & \textbf{0.48} \\ 
 & 1 & 0.28 & 0.37 & 0.09 & 0.24 & 0.1 & 0.25 & 0.04 & 0.43 & 0.09 & 0.08 & \textbf{0.57} \\ 
 & 5 & 0.28 & 0.44 & 0.1 & 0.32 & 0.16 & 0.21 & 0.09 & 0.54 & 0.08 & 0.06 & \textbf{0.6} \\ 
 & 10 & 0.31 & 0.49 & 0.09 & 0.4 & 0.17 & 0.28 & 0.05 & 0.54 & 0.09 & 0.06 & \textbf{0.64} \\ 
\multirow{4}{3em}{present-past}  & 0 & 0.15 & 0.31 & 0.04 & 0.11 & 0.04 & 0.1 & 0.04 & \textbf{0.42} & 0.04 & 0.04 & 0.39 \\ 
 & 1 & 0.22 & 0.4 & 0.05 & 0.16 & 0.21 & 0.22 & 0.05 & 0.52 & 0.05 & 0.05 & \textbf{0.6} \\ 
 & 5 & 0.23 & 0.41 & 0.05 & 0.13 & 0.15 & 0.25 & 0.05 & 0.59 & 0.14 & 0.05 & \textbf{0.62} \\ 
 & 10 & 0.23 & 0.41 & 0.05 & 0.15 & 0.2 & 0.49 & 0.05 & 0.52 & 0.14 & 0.05 & \textbf{0.66} \\ 
\multirow{4}{3em}{product-by-company}  & 0 & 0.15 & 0.23 & 0.07 & 0.21 & 0.03 & \textbf{0.51} & 0.03 & 0.32 & 0.06 & 0.07 & 0.31 \\ 
 & 1 & 0.22 & 0.35 & 0.07 & 0.15 & 0.09 & 0.16 & 0.05 & \textbf{0.43} & 0.08 & 0.08 & 0.41 \\ 
 & 5 & 0.27 & 0.41 & 0.1 & 0.33 & 0.1 & 0.16 & 0.05 & 0.48 & 0.1 & 0.1 & \textbf{0.49} \\ 
 & 10 & 0.28 & 0.42 & 0.1 & 0.29 & 0.12 & 0.19 & 0.05 & 0.43 & 0.1 & 0.1 & \textbf{0.51} \\
 
    \bottomrule
  \end{tabular}
  \end{small}
  \caption{Llama2-7B perturbation test on tasks of natural questions.}
  \label{tab: perturbation appendix llama2-7B natural}
\end{table*}

\newpage

\clearpage
\clearpage

\begin{figure*}
  \centering
  \includegraphics[width=0.9\linewidth]{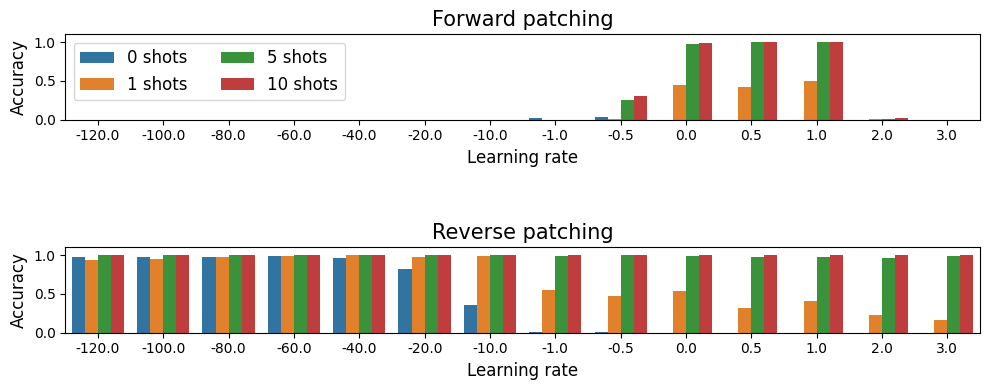}
  \caption{Forward and Reversed attention patching results are presented as a function of the learning rate, a scalar used in the injection of attention maps. The provided results are for GPT2-xl and the capitalized ICL task, demonstrating that Reversed patching can achieve the same results as prompting the model with 10-shots, even with a 0-shot prompt.}
\label{fig: patching lr}
\end{figure*}

\section{Attention Patching}
\label{Attention patching}
In this section, we provide additional information about the implementation of attention patching.

\paragraph{Learning rate}
Attention patching involves adding (injecting) attention maps into the forward pass of a given model.
The attention maps that are injected can be taken from the forward pass or from the RA, each time after being multiplied by a learning rate.
\autoref{fig: patching lr} demonstrates the effect of different learning rates on the success of attention patching. 
If we inject forward pass maps, a successful learning rate is a positive number.
For reversed attention, a negative learning rate would improve the model.
In our experiments, we used a learning rate of $1$ for the forward pass attention, and for the reversed attention, we used a learning rate of $-30$ for all models. Although we found that different learning rates might affect the models' performances, we decided not to tune the learning rate per task or model since our goal is to show proof of concept rather than optimizing the model accuracy in our experiments.

\paragraph{Tasks}
The bottleneck of attention patching is the need to have the same length and format of examples.
We use the ICL tasks from \autoref{Datasets} to construct sets of examples with the same length.
For each task, we randomly pick one of those sets and use it in our experiments. Following the perturbation test \autoref{Perturbation test}, we used 25 examples to average the forward or reversed attention maps for each attention head.

\paragraph{Additional results and discussion}
The full patching results are presented in \autoref{tab: patching appendix} .

The results show that RA patching can achieve similar results to ICL prompting, without the need to provide examples for the model to learn from.
Surprisingly, forward attention also shows improvement over the model's original accuracy, but not a consistent one across all tasks.
One possible reason for this is the ICL setup, in which the model does not identify the task it needs to perform, and hence returns irrelevant answers.
When averaging the forward pass attention scores, it might aggregate multiple different forward passes, amplifying common patterns and reducing noise attribute to the lack of context in the model's prompt.
In this sense, forward attention patching can be seen as a form of regularization or a compression of multiple examples' attention scores into one.

To the best of our knowledge, intervention on the attention maps has yet to be explored directly as our implementation. We used RA as a method to inject the attention maps we suspect would improve the model's performance. The injection of the forward pass, despite its surprisingly partial success, was added as a baseline to the RA patching. Future work might explore more complex methods of injecting attention maps, potentially overcoming the need for a constant size of attention maps during the patching process.

\section{Additional Implementation Details}
\paragraph{Compute:} All experiments were conducted on standard Nvidia-A40 series GPU.

\paragraph{Models:} All LM models were accessed via HuggingFace transformers stack \cite{wolf2019huggingface}.

\begin{table*}
  \centering
  \begin{small}
  \begin{tabular}{@{}l@{~~}l@{~~}c@{~~}c@{~~}c@{~~}c@{~~}c@{~~}c@{~~~~}|l@{~~}l@{~~}c@{~~}c@{~~}c@{~~}c@{~~}c@{~~}c@{~~}c@{}}
    \toprule
        & &  \multicolumn{3}{c}{GPT2-xl} & \multicolumn{3}{c|}{OPT-1.3B} &   & \multicolumn{3}{c}{GPT2-xl} & \multicolumn{3}{c}{OPT-1.3B}                    \\
     \cmidrule(lr){3-5}
     \cmidrule(lr){6-8}
        \cmidrule(lr){11-13}
     \cmidrule(lr){14-16}

     Task & N & original & FA & RA &
                original & FA & RA & 
    Task & N & original & FA & RA & 
                original & FA & RA \\
\midrule 
        \multirow{4}{3.5em}{adjective-v-verb-3} & 0 & 0.47 & 0.51 & \textbf{1.00} & 0.22 & 0.61 & \textbf{0.96}  &         \multirow{4}{3.5em}{animal-v-object-3} & 0 & 0.36 & 0.39 & \textbf{1.00} & 0.18 & 0.21 & \textbf{1.00} \\ 
       & 1 & 0.94 & \textbf{1.00} & \textbf{1.00} & 0.94 & 0.80 & \textbf{1.00}  &        & 1 & 0.96 & \textbf{1.00} & \textbf{1.00} & 0.82 & 0.96 & \textbf{1.00} \\ 
       & 5 & \textbf{1.00} & 0.90 & \textbf{1.00} & \textbf{1.00} & \textbf{1.00} & \textbf{1.00}  &        & 5 & \textbf{1.00} & 0.21 & \textbf{1.00} & \textbf{1.00} & \textbf{1.00} & \textbf{1.00} \\ 
       & 10 & \textbf{1.00} & 0.00 & \textbf{1.00} & \textbf{1.00} & \textbf{1.00} & \textbf{1.00}  &        & 10 & 0.96 & 0.00 & \textbf{1.00} & \textbf{1.00} & \textbf{1.00} & \textbf{1.00} \\ 
\midrule 
        \multirow{4}{3.5em}{antonym} & 0 & 0.00 & 0.01 & \textbf{0.08} & 0.02 & 0.01 & \textbf{0.24}  &         \multirow{4}{3.5em}{capitalize} & 0 & 0.00 & 0.00 & \textbf{0.94} & 0.01 & 0.00 & \textbf{0.78} \\ 
       & 1 & 0.18 & 0.43 & \textbf{0.56} & 0.20 & 0.26 & \textbf{0.57}  &        & 1 & 0.44 & 0.50 & \textbf{1.00} & 0.01 & 0.01 & \textbf{0.90} \\ 
       & 5 & 0.53 & 0.57 & \textbf{0.62} & 0.42 & 0.44 & \textbf{0.59}  &        & 5 & 0.98 & \textbf{1.00} & \textbf{1.00} & \textbf{1.00} & 0.99 & \textbf{1.00} \\ 
       & 10 & 0.57 & 0.57 & \textbf{0.62} & 0.42 & 0.43 & \textbf{0.54}  &        & 10 & 0.99 & \textbf{1.00} & \textbf{1.00} & \textbf{1.00} & 0.99 & \textbf{1.00} \\ 
\midrule 
        \multirow{4}{3.5em}{choose-middle-of-3} & 0 & 0.46 & 0.30 & \textbf{1.00} & 0.11 & 0.03 & \textbf{1.00}  &         \multirow{4}{3.5em}{english-french} & 0 & 0.00 & 0.00 & 0.00 & 0.00 & 0.00 & 0.00 \\ 
       & 1 & 0.81 & 0.76 & \textbf{1.00} & 0.57 & 0.54 & \textbf{0.76}  &        & 1 & 0.00 & 0.00 & 0.00 & 0.00 & 0.00 & \textbf{0.01} \\ 
       & 5 & 0.92 & 0.68 & \textbf{1.00} & 0.95 & 0.84 & \textbf{1.00}  &        & 5 & 0.01 & 0.01 & \textbf{0.03} & 0.00 & 0.00 & \textbf{0.04} \\ 
       & 10 & 0.97 & 0.00 & \textbf{1.00} & 0.86 & 0.65 & \textbf{1.00}  &        & 10 & 0.01 & 0.01 & \textbf{0.03} & \textbf{0.03} & 0.00 & \textbf{0.03} \\ 
\midrule 
        \multirow{4}{3.5em}{landmark-country} & 0 & 0.00 & 0.00 & 0.00 & 0.00 & 0.00 & 0.00  &         \multirow{4}{3.5em}{next-capital-letter} & 0 & \textbf{0.01} & \textbf{0.01} & \textbf{0.01} & \textbf{0.01} & 0.00 & \textbf{0.01} \\ 
       & 1 & 0.31 & 0.22 & \textbf{0.39} & 0.25 & 0.33 & \textbf{0.47}  &        & 1 & 0.03 & 0.03 & \textbf{0.05} & \textbf{0.05} & 0.02 & 0.02 \\ 
       & 5 & \textbf{0.47} & 0.39 & 0.44 & \textbf{0.53} & 0.42 & 0.47  &        & 5 & 0.05 & 0.02 & \textbf{0.06} & \textbf{0.03} & \textbf{0.03} & 0.01 \\ 
       & 10 & 0.44 & \textbf{0.47} & 0.44 & \textbf{0.50} & 0.33 & 0.42  &        & 10 & 0.04 & \textbf{0.06} & \textbf{0.06} & 0.04 & 0.03 & \textbf{0.08} \\ 
\midrule 
        \multirow{4}{3.5em}{next-item} & 0 & 0.03 & 0.00 & \textbf{0.16} & 0.09 & 0.00 & \textbf{0.53}  &         \multirow{4}{3.5em}{present-past} & 0 & 0.03 & \textbf{0.05} & \textbf{0.05} & 0.05 & 0.05 & \textbf{0.07} \\ 
       & 1 & 0.28 & 0.66 & \textbf{0.72} & 0.50 & 0.47 & \textbf{0.84}  &        & 1 & 0.37 & 0.72 & \textbf{0.88} & \textbf{0.13} & \textbf{0.13} & \textbf{0.13} \\ 
       & 5 & 0.69 & 0.84 & \textbf{0.88} & 0.69 & 0.66 & \textbf{0.88}  &        & 5 & 0.95 & 0.97 & \textbf{0.98} & 0.80 & 0.58 & \textbf{0.97} \\ 
       & 10 & 0.88 & 0.88 & \textbf{0.91} & 0.75 & 0.81 & \textbf{0.84}  &        & 10 & 0.92 & \textbf{0.97} & 0.93 & \textbf{0.98} & 0.67 & 0.95 \\ 
\midrule 
        \multirow{4}{3.5em}{prev-item} & 0 & 0.03 & 0.06 & \textbf{0.19} & 0.03 & 0.03 & \textbf{0.06}  &         \multirow{4}{3.5em}{singular-plural} & 0 & 0.00 & \textbf{0.09} & \textbf{0.09} & \textbf{0.09} & \textbf{0.09} & \textbf{0.09} \\ 
       & 1 & 0.25 & \textbf{0.34} & 0.25 & \textbf{0.09} & 0.06 & \textbf{0.09}  &        & 1 & 0.52 & 0.70 & \textbf{0.91} & 0.43 & 0.78 & \textbf{0.96} \\ 
       & 5 & \textbf{0.50} & 0.41 & 0.47 & 0.44 & 0.31 & \textbf{0.53}  &        & 5 & \textbf{1.00} & \textbf{1.00} & 0.91 & 0.87 & \textbf{1.00} & 0.96 \\ 
       & 10 & 0.50 & \textbf{0.59} & 0.50 & 0.47 & 0.38 & \textbf{0.66}  &        & 10 & \textbf{1.00} & \textbf{1.00} & \textbf{1.00} & 0.96 & \textbf{1.00} & 0.96 \\ 
\midrule 
        \multirow{4}{3.5em}{synonym} & 0 & 0.01 & 0.01 & \textbf{0.02} & 0.01 & 0.01 & \textbf{0.02}  &  \\ 
       & 1 & 0.03 & 0.04 & \textbf{0.05} & 0.04 & 0.05 & \textbf{0.13}  &  \\ 
       & 5 & 0.04 & 0.06 & \textbf{0.10} & 0.06 & 0.02 & \textbf{0.18}  &  \\ 
       & 10 & 0.03 & 0.06 & \textbf{0.10} & 0.06 & 0.02 & \textbf{0.23}   \\

    \bottomrule
  \end{tabular}
  \end{small}
  \caption{GPT2-xl and OPT-1.3B accuracy on ICL tasks with forward attention (FA) and Reversed Attention (RA) patching.}
  \label{tab: patching appendix}
\end{table*}

\end{document}